\documentclass[fleqn,10pt]{wlpeerj}

\usepackage{amssymb}
\usepackage{amsmath}

\usepackage{graphicx}

\usepackage{multirow}%
\usepackage{amsmath,amssymb,amsfonts}%
\usepackage{amsthm}%
\usepackage{mathrsfs}%
\usepackage[title]{appendix}%
\usepackage{xcolor}%
\usepackage{textcomp}%
\usepackage{manyfoot}%
\usepackage{booktabs}%
\usepackage{algorithm}%
\usepackage{algorithmicx}%
\usepackage{algpseudocode}%
\usepackage{listings}%
\usepackage{rotating} %
\usepackage{subcaption}%
\usepackage{comment}%
\usepackage[hidelinks]{hyperref}

\def\Rottt#1{ \multicolumn{1}{l||}{\rlap{\rotatebox{90}{#1}~}}}
\def\Rot#1{ \multicolumn{1}{l|}{\rlap{\rotatebox{90}{#1}~}}}
\def\Rote#1{ \multicolumn{1}{l}{\rlap{\rotatebox{90}{#1}~}}}

\theoremstyle{thmstyleone}%
\theoremstyle{thmstyletwo}%

\theoremstyle{thmstylethree}%

\newcommand\cat[1]{\texttt{#1}}

\newcommand\change[1]{\colorlet{saved}{.}\color{blue}{#1 }\color{saved}}

\title{Modelling and Classifying the Components of a Literature Review}

\author[1]{Francisco Bolaños}
\author[1]{Angelo Salatino}
\author[1,2]{Francesco Osborne}
\author[1]{Enrico Motta}
\affil[1]{Knowledge Media Institute, The Open University, Walton Hall, Milton Keynes, MK7 6AA, UK}

\affil[2]{Department of Business and Law, University of Milano Bicocca, Piazza dell'Ateneo Nuovo, 1, Milan, 20126, IT}

\corrauthor[1]{Francisco Bolaños}{francisco.bolanos-burgos@open.ac.uk}

\begin{abstract}
Previous work has demonstrated that AI methods for analysing scientific literature benefit significantly from annotating sentences in papers according to their rhetorical roles, such as research gaps, results, limitations, extensions of existing methodologies, and others. Such representations also have the potential to support the development of a new generation of systems capable of producing high-quality literature reviews. However, achieving this goal requires the definition of a relevant annotation schema and effective strategies for large-scale annotation of the literature. This paper addresses these challenges in two ways: 1) it introduces a novel, unambiguous annotation schema that is explicitly designed for reliable automatic processing, and 2) it presents a comprehensive evaluation of a wide range of large language models (LLMs) on the task of classifying rhetorical roles according to this schema. To this end, we also present \textit{Sci-Sentence}, a novel multidisciplinary benchmark comprising 700 sentences manually annotated by domain experts and 2,240 sentences automatically labelled using LLMs. We evaluate 37 LLMs on this benchmark, spanning diverse model families and sizes, using both zero-shot learning and fine-tuning approaches. The experiments reveal that modern LLMs achieve strong results on this task when fine-tuned on high-quality data, surpassing 96\% F1, with both large proprietary models such as GPT-4o and lightweight open-source alternatives performing well. Moreover, augmenting the training set with semi-synthetic LLM-generated examples further boosts performance, enabling small encoders to achieve robust results and substantially improving several open decoder models.
\end{abstract}

\begin{document}

\flushbottom
\maketitle
\thispagestyle{empty}

\section{Introduction}\label{sec:introduction}

A literature review or related work section is a fundamental component of a research paper, as it provides the necessary background, highlights the research gap, and justifies the research objectives~\citep{booth2021systematic}. It also serves to summarise relevant literature in educational settings, aiding students and researchers in understanding the state of the art regarding a certain topic. However, crafting a high-quality literature review remains a challenging task, even for experienced researchers. It requires comprehensive knowledge of the relevant literature, which is increasingly difficult to maintain due to the growing volume of published research~\citep{bornmann2015} and the continual need for updates to ensure relevance~\citep{moher2007systematic}. In addition, it demands the ability to synthesise this information into a clear and structured discussion that highlights key research directions, theoretical frameworks, and open challenges in the field.

In recent years, artificial intelligence (AI) and natural language processing (NLP) research has focused on analysis and automatic generation of scientific literature.
Previous studies~\citep{wang2022representation,groza2013using,widyantoro2013multiclass,Teufel2002ArticlesSS} have shown that AI methods for analyzing scientific literature benefit significantly from annotating sentences in papers according to their rhetorical roles, such as research gaps, results, and limitations. Texts annotated with such roles have been demonstrated to facilitate the analysis of the evolution of scientific knowledge~\citep{wang2022representation} and to assist in identifying~\citep{groza2013using,widyantoro2013multiclass} and predicting~\citep{Teufel2002ArticlesSS} the significance of scientific concepts and contributions.

More recently, the rise of large language models (LLMs) has led to the emergence of various systems that claim to generate scientific analysis by leveraging relevant scholarly literature~\citep{shi2023towards,zhang2024references,li2024explaining,martin2024shallow,zhang2024litfm,nishimura2024toward,li2024chatcite,ma2025refinement,agarwal2025litllms,liu2025select,beger2025citegeist,wang2025scholarcopilot}.
Many of these systems focus on the specific task of related work generation, which involves producing coherent and informative text that synthesises and discusses relevant aspects of the existing literature, similar to the content typically found in the literature review or related work sections of scientific papers. 
However, despite the surface coherence of the outputs produced by current LLM-based approaches, the quality of the resulting literature reviews remains limited. This limitation primarily arises because these outputs often consist of uncritical summaries of individual papers rather than structured and analytical reviews that take into consideration important concepts such as research gaps and  limitations~\citep{bolanos2024artificial}.
To support the development of systems capable of producing high-quality literature reviews, we argue that it is essential to integrate rhetorical frameworks for literature analysis which identify the functional role of each sentence with the natural language generation capabilities of LLMs. Such integration would enable systems to leverage a richer representation of the input text, allowing, for example, the retrieval of all sentences that discuss research gaps within a specific domain in order to generate coherent overviews focused on future research directions.

The first step toward achieving this integration is to address the task of classifying text according to its rhetorical structure, thereby producing a representation that can support downstream systems for literature analysis and evaluation. In this study, we focus particularly on recent transformer-based architectures, which have demonstrated strong performance in related tasks involving the classification of technical text~\citep{fields2024survey}. 
This objective gives rise to two main research questions.
First, what kind of annotation schema can enable a fully automatic characterisation of sentences in literature reviews?
Second, how can current NLP technologies, particularly transformer models, reliably identify the rhetorical roles of sentences in research papers?

This paper tackles these challenges by: 1) introducing a novel annotation schema explicitly designed to support reliable automatic processing, and 2) performing a comprehensive evaluation of state-of-the-art transformer models on a human-curated benchmark constructed according to this schema. 

As first step, we developed an annotation schema inspired by prior studies on rhetorical structure~\citep{khoo2011analysis, jaidka2013literature}, which categorises scientific sentences into seven classes: \cat{Overall}, \cat{Research Gap}, \cat{Description}, \cat{Result}, \cat{Limitation}, \cat{Extension}, and \cat{Other}. 
Unlike earlier approaches~\citep{khoo2011analysis}, which were primarily developed for manual annotation and used a few potentially ambiguous labels requiring contextual interpretation beyond the capability of automated classifiers, our schema defines clear and unambiguous categories that can be consistently annotated by machine learning models.

Based on this schema, we created a new publicly available resource, \textit{Sci-Sentence}, a multidisciplinary benchmark that includes 700 sentences manually annotated by domain experts and 2,240 sentences automatically labelled using LLMs. 
We then evaluated several categories of transformer models on this dataset, covering encoder-only, encoder-decoder, and decoder-only architectures. The evaluation compared models of different sizes (1–3B, 7–8B, and large-scale LLMs), various prominent model families (e.g., BERT, T5, Llama, Mistral, Gemma, Phi), and both proprietary and open-source variants. In total, we tested 37 models, organised into distinct categories, using both zero-shot and fine-tuning settings. The aim was to assess the overall performance of transformer-based models and to gain deeper insights into the most effective approaches and optimisation strategies for this challenging task. 
We also conducted an in-depth error analysis to identify the categories that pose the greatest challenges for automatic recognition across different model types. For example, we found that the \cat{Limitation} and \cat{Description} categories are particularly difficult, as they can sometimes appear in forms that may be misconstrued as other types. Furthermore, we compared two fine-tuning optimisation techniques, LoRA and NEFT, and observed that LoRA achieved superior results for large decoder models. Finally, we investigated the impact of training on synthetic data across different model architectures. 

The experiments yielded several novel insights that advance the state of the art in this challenging domain.
First, the current generation of LLMs performs remarkably well on this task when fine-tuned on high-quality datasets such as \textit{Sci-Sentence}, reaching performance levels above 96\% F1.
Second, while large proprietary models like GPT-4o achieve the best results, lightweight open-source alternatives, such as \textit{SuperNova-Medius} and \textit{Nemotron-8B}, also demonstrate excellent performance.
Third, although decoder-only models achieve the highest overall scores, small and scalable encoder-based models pre-trained on domain-relevant data, such as \textit{SciBERT}, also achieve solid performance. Therefore, they represent a practical solution for efficiently processing large volumes of text.
Finally, enriching the training data with semi-synthetic examples generated by LLMs has proven very beneficial. This solution enables small encoders to achieve robust results and significantly enhances the performance of several open decoder models.

The remainder of this paper is structured as follows. Section~\ref{sec:related work} reviews related work, including established literature review frameworks and existing approaches for classifying sentences in scientific articles. Section~\ref{sec:background} defines the task, describes the development of the annotation schema, and presents the novel benchmarks. Section~\ref{sec:methodology} details the experimental methodology and describes the models and approaches used to classify scientific sentences. Section~\ref{sec:results} presents the experimental results, with additional analysis on the performance of different methods, optimization techniques, and the effectiveness of semi-synthetic data. Finally, Section~\ref{sec:conlusion} concludes the paper and outlines potential directions for future research.

\section{Related Work}\label{sec:related work}

We review the current literature by focusing on two main research strands. First, we examine existing frameworks for categorising the content of a research paper based on rhetorical roles and discourse analysis (Section~\ref{subsec:lr_framework}). Second, we survey various NLP approaches for classifying scientific sentences in the context of generating related work sections (Section~\ref{subsec:llm_classification}).



\subsection{Frameworks based on the Rhetorical Structure of Scientific Papers}\label{subsec:lr_framework}


Rhetorical Structure Theory is a framework for text organisation that has inspired applications in discourse analysis, text generation, psycholinguistics, and computational linguistics~\citep{taboada2006applications}. It has also been extensively applied to the study, understanding, and generation of scientific and scholarly texts. In particular, three main types of analyses have emerged from its application to scientific and academic literature: \textit{genre analysis}~\citep{swales1990genre, kanoksilapatham2005rhetorical}, \textit{zoning analysis}~\citep{teufel1999annotation, mizuta2006zone, teufel2009towards, Liakata2010CorporaFT}, and \textit{discourse analysis}~\citep{mann1988rhetorical, Prasad2008ThePD}.


Genre analysis is commonly employed in the field of linguistics due to its effectiveness in manual rhetorical analysis and its pedagogical value in academic writing instruction~\citep{swales1990genre,swales2004academic,kwan2006schematic,bastola2023rhetorical}. It provides a structured guide to analysing and creating introductions of scientific papers or review chapters of doctoral theses. In contrast, zoning and discourse analyses are studied in NLP research, as they provide machine-interpretable categories and enable fine-grained, sentence-level annotation throughout the entire scholarly document. While zoning analysis focuses on identifying the rhetorical function of individual sentences within the scientific argument~\citep{teufel1999annotation,teufel2009towards}, discourse analysis examines the textual coherence, meaning, and structural relationships at the sentence level~\citep{khoo2011analysis}.

In the domain of related work generation
only a limited number of studies have investigated genre or discourse analysis.
\cite{wang2024can1}, inspired by the genre analysis, proposed the CaRS model, which describes how academic writers structure introductions by establishing, justifying, and presenting their work. The authors also introduced RSGen, a transformer-based model that employs a two-step decoding process: first creating a rhetorical plan, then generating the related work content. However, RSGen achieves only moderate performance and suffers from issues such as error propagation in classification and limited generalizability to alternative rhetorical schemas.
\cite{khoo2011analysis}, drawing on the discourse-based analysis, focused instead on the foundational task of understanding human-written literature reviews through manual analysis. They examined the macro-level discourse structure of literature reviews in information science journals, developing a coding schema with 12 categories.  
Nevertheless, the coding schema presents several challenges that hinder its suitability for fully automated adoption. First, some categories are defined ambiguously and can be easily misinterpreted even by human annotators, often requiring contextual knowledge that an automated classifier would not possess. Second, certain categories are overly broad, which may lead to confusion and reduced consistency in downstream systems. Finally, the schema was not originally designed for computational processing and therefore lacks reliable training datasets and clear, machine-readable classification guidelines.

In this study, we adopt a discourse-based approach inspired by the findings of Jaidka, Khoo \& Na~\citep{jaidka2013literature}, which emphasizes the critical role of sentences in the generation of related work section and recognizes the lack of a well-defined structure in existing research~\citep{shi2023towards,zhang2024references,li2024explaining,martin2024shallow,zhang2024litfm,nishimura2024toward,li2024chatcite,ma2025refinement,agarwal2025litllms,liu2025select,beger2025citegeist,wang2025scholarcopilot}. In particular, we seek to enhance the coding schema introduced by \cite{khoo2011analysis}, with the objective of formulating machine-interpretable categories that facilitate their automation by AI systems.

\subsection{Approaches for Classifying Sentences in Related Work Section Generation }\label{subsec:llm_classification}


The literature presents a variety of approaches for classifying sentences in scholarly articles. Several methods focus on the rhetorical structure of the paper~\citep{khoo2011analysis}, most often targeting the discourse around citations, including citation function~\citep{garfield1965can,dong2011ensemble,teufel2006automatic,jurgens2018measuring,tuarob2019automatic,zhao2019context}, citation intent~\citep{cohan2019structural,lauscher2021multicite}, and citation sentiment~\citep{athar2011sentiment,athar2012context,ravi2018article}. Other approaches aim to associate sentences or text segments with relevant topics~\citep{salatino2022cso,masoumi2023fuzzy}, often selected from one of the many knowledge organisation systems used to categorise scientific literature~\citep{salatino2025survey}.
Another category of systems focuses on extracting research entities (e.g., tasks, methods, materials) linked by semantic relations~\citep{dessi2022scicero}. The richer representations of the literature produced by these systems are often encoded in knowledge graphs~\citep{peng2023knowledge} (e.g., SemOpenAlex~\citep{farber2023semopenalex}, ORKG~\citep{jaradeh2019open}, AI-KG~\citep{dessi2020ai}, CS-KG~\citep{dessi2022cs}, Nano-publications~\citep{kuhn2016decentralized}) and have been shown to support effectively scientometric analyses~\citep{salatino2020,salatino2023}, intelligent systems for exploring the literature~\citep{angioni2021aida}, and, increasingly, conversational systems~\citep{meloni2023integrating} and question-answering methods~\citep{SciQA2023,lehmann2024large,DBLP:conf/birws/BanerjeeAUB23}.
However, as noted by several recent studies~\citep{li2022automatic,li2024related,bolanos2024artificial}, the output of existing approaches does not adequately support the generation of related work sections. Therefore, in this paper, we propose a novel sentence classification approach explicitly designed to support related work analysis and generation.

To the best of our knowledge, only two approaches for related work generation incorporate sentence classification into their pipeline~\citep{xing2020automatic,ge2021baco}, and both specifically focus on characterising citations. 
Before discussing these methods, it is useful to first introduce the strategies employed in related work generation. These strategies are commonly classified as either extractive or abstractive.  Extractive systems utilise sentences as their units of analysis, producing a paragraph by concatenating selected sentences in a specific order~\citep{hoang2010towards,hu2014automatic,wang2019neural,chen2019automatic,wang2019toc,deng2021automatic}. The paragraph does not have any division and lacks the structure of a human-written literature review~\citep{li2024related}. In contrast, abstractive systems process input such as excerpts or paragraphs generating fluent paragraphs~\citep{abura2020automatic,xing2020automatic,ge2021baco,chen2021capturing,luu2020explaining,jung2022intent,li2022corwa,chen2022target,li2023cited,liu2023causal,gu2024controllable,li2024explaining,martin2024shallow}. However, their outputs often exhibit deficiencies, such as the absence of transitional sentences, improper citation ordering, or as in extractive approaches the lack of structure~\citep{li2024related}.

The two approaches that use sentence classification are both abstractive methods. \cite{xing2020automatic} trained a BERT~\citep{kenton2019bert} model to classify sentences as explicit citations, which directly name the source, or implicit citations, which refer to the work without naming it. \cite{ge2021baco} fine-tuned SciBERT~\citep{beltagy2019scibert} to categorise citation sentences as positive, negative, or neutral, depending on whether they emphasise contributions, highlight shortcomings, or provide objective descriptions of the cited work.

In this paper, we significantly advance the state of the art in this domain by 1) proposing a new classification schema designed to support systems for related work generation and 2) exploring how LLMs can be used to automatically label sentences at scale.

\section{Framework}\label{sec:background}
This section presents and justifies the theoretical framework and the dataset used for the annotation of research papers to support the analysis and generation of related work.  
We begin by formally defining our task and outlining the categories (Section~\ref{sec:task_definition}).  
Next, we describe in detail the development of the Sci-Sentence Benchmark (Section~\ref{sec:benchmark}).



\subsection{Task definition and Annotation Schema}\label{sec:task_definition}\label{sec:categories}


To support automatic systems in creating more structured literature reviews, we will characterize text from the related work sections into specific rhetorical types. Specifically, we propose classifying each sentence in these sections into a single, specific rhetorical type.
We frame this as a single-label multi-class classification problem, where each sentence is assigned to the most appropriate category.
The annotation schema presented in this paper includes seven categories: \cat{Overall}, \cat{Research Gap}, \cat{Description}, \cat{Result}, \cat{Limitation}, \cat{Extension}, and \cat{Other}. These categories are defined in Table~\ref{tab:AnnotationSchema}. 




\begin{table}[!h]
  \centering
  \footnotesize
  \caption{Proposed annotation schema.}
  \label{tab:AnnotationSchema}
  \begin{tabular}{l|p{7.5cm}|l}
    \toprule
    \textbf{Category} & \textbf{Description}& \textbf{Association} \\
    \midrule
    Overall & It is a sentence describing, introducing, classifying, or defining a  research topic often based on the discussion of multiple previous studies together.&Topic \\ \hline
    Research Gap & It is a sentence highlighting the need for new research in a topic due to  absence of information, insufficient information or  contradictory information.&Topic  \\\hline
    Description & It is a sentence describing the objective, methodology or design of a previous study.& Study  \\ \hline
    Result & It is a sentence presenting the findings of a previous study.& Study  \\ \hline
    Limitation & It is a sentence describing any factor that can affect the validity or reliability of the previous study regarding its methodology. & Study  \\  \hline   
    Extension & It is a sentence describing how the current study addresses or extends previous studies by stating the overall idea, contrasting ideas or elaborating further ideas. & Study\ \\  \hline
    Other & This denotes a sentence that does not fit within the above categories.& None \\ \bottomrule
  \end{tabular}
\end{table}

This annotation schema was developed by building on the theoretical work of \cite{khoo2011analysis}, who proposed an influential coding schema for annotating the macro-level discourse structure of related work sections.
However, the original schema was not optimised for automated processing and therefore requires revision to ensure its suitability for machine learning classifiers.
Some categories have soft boundaries that partially overlap, and their definitions depend on substantial contextual understanding, which hinders both consistent human annotation and automated interpretation. In addition, several categories are overly broad, introducing potential ambiguity for downstream automatic systems that rely on precise classification. For example, the ``what'' category may refer either to a general concept in the field or to a specific methodology introduced in the same paper.




To address these issues, we followed the protocol of \cite{khoo2011analysis} and conducted an iterative annotation process on over 300 sentences extracted from research papers in Computer Science, Business, Medicine, and Psychology. We initially applied their original coding schema and then systematically refined it by merging overlapping categories and splitting ambiguous ones. This process resulted in a new set of distinct and clearly defined classifications that are more suitable for consistent interpretation by AI systems.

The main change concerns the introduction of a clear distinction between i) sentences that describe the overall research topic and ii) sentences that refer to individual studies. These two levels are not explicitly addressed in the original framework by \cite{khoo2011analysis}, yet they are practically important for supporting both human and automated annotation. For example, the categories \textit{what}, \textit{description}, and \textit{method} in the original schema can apply to either the topic level or the study level, and their function varies significantly depending on the context.

In our classification schema, each category is restricted to one of these two levels in order to avoid ambiguity. The topic level is represented by the categories \cat{Overall}, which provides a general overview of the research area, and \cat{Research Gap}, which highlights unresolved issues or open questions in the field. The study level includes four categories that are specific to individual studies: \cat{Description}, \cat{Result}, \cat{Limitation}, and \cat{Extension}.

We also introduced the category \cat{Other} because an automatic classifier sometimes encounters sentences that do not match any predefined category. This additional category prevents the model from assigning an incorrect label when its confidence is low or when the text is unrelated to the existing categories.

\begin{table}[!h]
  \centering
  \footnotesize
  \caption{Comparison between the two coding schemas.}
  \label{tab:AnnotationSchemaComparison}
  \begin{tabular}{l|l|l}
    \toprule
    \textbf{Proposed schema} & \textbf{Association} & \textbf{Khoo et al.~\cite{khoo2011analysis} schema} \\
    \midrule
    Overall & Topic & What, Description, Meta‐Summary, Brief‐Topics \\
    Research Gap & Topic & Meta‐critique\\
    Description & Study &What, Description, Method \\
    Result & Study & Result\\
    Limitation & Study & Meta‐Critique\\   
    Extension & Study & Current‐Study\\
    Other & -& -\\

    \bottomrule
  \end{tabular}
\end{table}

Table~\ref{tab:AnnotationSchemaComparison} compares the coding schema proposed by \cite{khoo2011analysis} with our revised schema. 
Notably, we deconstruct Khoo et al.'s broad \textit{meta-critique} category, which may conflate critiques of either a topic or a specific study, into two categories: \cat{Research Gap}, which signals the need for further research at the topic level, and \cat{Limitation}, which identifies methodological or conceptual shortcomings in a particular study.
Our schema also resolves the ambiguity in Khoo et al.'s interchangeable use of \textit{what} and \textit{description} across both levels. We use \cat{Overall} exclusively for sentences that describe the research area as a whole, and reserve \cat{Description} for sentences detailing individual studies. The \cat{Overall} category also consolidates several of Khoo et al.'s categories (namely \textit{what}, \textit{description}, \textit{meta-summary}, and \textit{brief-topics}) into a single, more coherent label for topic-level summaries. 
The \cat{Extension} category has also been reconceptualised. While Khoo et al. use \textit{current-study} for sentences that refer to the current work in a general manner and can therefore be easily conflated with \cat{Description} or \cat{Limitation}, our definition captures the motivations underlying the current work. This includes the articulation of new ideas, the identification of contrasting perspectives, and the elaboration of existing approaches.

\subsection{The Sci-Sentence Benchmark}\label{sec:benchmark}

To evaluate the ability of modern LLMs to classify sentences according to the annotation schema introduced earlier, we developed the \textit{Sci-Sentence Benchmark}. \textit{Sci-Sentence} includes 700 sentences manually annotated by domain experts, along with 2,240 automatically labelled sentences. 
These sentences were extracted from the introduction, related work, and limitations sections of scientific papers in Computer Science, Business, Education, Medicine, and Psychology.



Sci-Sentence was developed in three phases. First, we conducted a workshop involving domain experts to define the annotation guidelines and compute inter-annotator agreement on a sample of 140 sentences. Second, the same experts individually annotated an additional 560 sentences, resulting in a total of 700 manually annotated sentences. This process produced the first, fully manually annotated, version of Sci-Sentence, which included 560 sentences for training and validation, and 140 sentences for testing. 
Finally, we leveraged Sonnet 3.0 to generate a larger version of the training and validation set by producing four alternative versions of each of the 560 sentences. This approach enabled the use of a more extensive training dataset for automatic methods, while retaining the fully manually annotated test set for evaluation purposes. 
In the following, we provide a more detailed account of the development process.

The three annotators who attended the workshop were researchers in Computer Science and Biology. To ensure consistency across annotations, a preliminary one-hour coordination session was held. The full annotation process took approximately three hours to complete. The resulting dataset included 140 sentences, selected from a larger pool of 300 annotated sentences, such that each of the seven categories was represented by 20 sentences. This sampling strategy was necessary because the categories \cat{Result} and \cat{Limitation} were infrequently observed in the original dataset and were therefore underrepresented.

To demonstrate the feasibility of the annotation task, the clarity of the category definitions, and the consistency of the expert annotations, we assessed inter-rater agreement. Specifically, we employed two metrics: Fleiss' Kappa~\citep{fleiss1971measuring} and Gwet’s AC1~\citep{gwet2008computing}. Fleiss' Kappa was used to measure the overall agreement among the three raters across the entire annotation exercise~\citep{fleiss1971measuring}. In contrast, Gwet’s AC1 was applied to evaluate agreement at the category level. This choice was motivated by the fact that Gwet’s AC1 is designed to overcome certain limitations of Fleiss' Kappa, providing a more robust and stable measure of inter-rater reliability under varying prevalence and marginal distribution conditions~\citep{gwet2008computing}.

A Fleiss’ Kappa of 0.90 was achieved for the overall agreement among the three raters, indicating a high level of inter-annotator reliability \citep{landis1977measurement}. Table~\ref{tab:Agreement} reports Gwet’s AC1 scores for category-specific agreement. In most cases, the agreement is above 0.80. The only exceptions are the \cat{Research Gap} and \cat{Limitation} categories, with agreement values of 0.78 and 0.75, respectively, which still represent substantial agreement according to established guidelines~\citep{landis1977measurement}. This strong general and category-specific agreement indicates that the annotation task is well-defined and that the experts were able to produce consistent labels. In turn, this suggests that the Sci-Sentence benchmark can serve as a high-quality resource for training downstream applications.


\begin{table}[!h]
\centering
\footnotesize
\caption{Average Gwet’s AC1 per Category}
\label{tab:Agreement}
\begin{tabular}{l|r}
\toprule
\textbf{Category}     & \textbf{Gwet's AC1}  \\
\midrule
Overall           & 0.89              \\
Research Gap      & 0.78              \\
Description       & 0.89              \\
Result            & 0.89              \\
Limitation        & 0.75              \\
Extension         & 0.93              \\
Other             & 0.97              \\ 
\bottomrule
\end{tabular}
\end{table}

In the second phase of developing the Sci-Sentence benchmark, the three annotators independently continued the annotation process, each working on a distinct set of sentences in accordance with the original guidelines. Annotation proceeded until each category was expanded by an additional 80 sentences, resulting in a total of 560 new annotated sentences. As a result, the complete dataset now contains 700 sentences, which are split into 70\% for training (490 sentences), 10\% for validation (70 sentences), and 20\% for testing (140 sentences). 

Given the labour-intensive and time-consuming nature of the annotation process, we also explored the use of semi-synthetic data. Unlike fully-synthetic samples, which are generated by mimicking the statistical properties of a dataset, semi-synthetic data refers to data generated considering the representation of real-world objects, such as original sentences~\citep{joshi2024synthetic}. Specifically, we aimed to generate artificial sentences that replicate the characteristics of those found in the original dataset~\citep{marwala2023use}. This approach has gained considerable momentum with the advent of LLMs~\citep{liu2024best}. 
The literature provides compelling evidence that semi-synthetic data can enhance dataset diversity~\citep{he2021semi}, support threats detection in security~\citep{MYNENI2023109688}, address missing values~\citep{berti2018discovery}, mitigate algorithmic bias~\citep{li2022more}, and support privacy-preserving data sharing~\citep{yale2019privacy}. In line with recent studies~\citep{kaddour2023synthetic,li2023synthetic,li2024data}, our experimental findings (see Section~\ref{sec:results}) confirm the effectiveness of this strategy.

To this end, we employed Sonnet 3.0 (accessed via Amazon Bedrock) to generate four semi-synthetic sentences for each original sentence in the training and validation sets only, leaving the test set unaltered~\citep{stan2025learningreasoningfailuressynthetic}. We subsequently evaluated the generated sentences to verify that they were sufficiently syntactically distinct from their corresponding source sentences. For this purpose, we used the normalised Levenshtein distance (see Appendix~\ref{secA}), which measures the similarity between two sentences on a scale from 0 (identical) to 1 (completely different). 
Specifically, we computed the normalised Levenshtein distance between each original sentence and its four new variants, as well as the average distance between each new sentence and the remaining three. If any of these distances fell below or equal to 0.20, indicating insufficient syntactic variation, these sentences were discarded and regenerated using the language model until they met the threshold.

Appendix~\ref{secA} includes the prompt used to generate the semi-synthetic sentences and a table reporting the normalised Levenshtein distances between sentences, grouped by category. This process resulted in 1,960 new sentences for the training set and 280 for the validation set, bringing the total number of sentences to 2,450 and 350, respectively, when combined with the manually annotated data.

We released the Sci-Sentence benchmark in two versions: (1) the \textbf{base version}, which contains 700 manually annotated sentences; and (2) the \textbf{augmented version}, which includes a total of 2,940 sentences comprising both the manually annotated data and additional semi-synthetic examples. As previously mentioned, in both versions the test set consists exclusively of manually annotated sentences to ensure a fair evaluation. The benchmark is available on GitHub under a CC-BY license\footnote{Datasets in  the Sci-Sentence Repository – \url{https://github.com/fcobolanos/Classifying-the-Components-of-a-Literature-Review/tree/main/datasets}}.  

\section{Experimental Methodology}\label{sec:methodology}

This paper aims to investigate the capability of AI models to annotate the rhetorical roles of sentences within research papers. It has two main objectives: i) to introduce a new annotation schema and a relevant dataset, detailed in Section~\ref{sec:background}; and ii) to examine whether current language models can accurately and efficiently perform this task at scale, as well as to identify which architectures are most effective.

This section describes the experimental methodology related to the second objective. Specifically, we conducted a comprehensive evaluation of a broad set of state-of-the-art LLMs~\citep{patel2023magnifico,kasneci2023chatgpt} on the novel Sci-Sentence Benchmark, under both zero-shot learning (ZSL) and fine-tuning settings. We assessed 37 alternative solutions spanning a variety of model architectures, including encoder-only, decoder-only, and encoder-decoder, and covering a wide range of parameter sizes. We also tested both open-source and proprietary solutions. 

All models were evaluated on the test set of the Sci-Sentence Benchmark, and their performance was measured using Precision, Recall, and F1-score.

The following subsections present the experiments conducted using ZSL (Section~\ref{sec:one-shot_method}) and fine-tuning (Section~\ref{sec:fine-tuning_method}). Finally, Section~\ref{sec:list of models} provides an overview of the employed LLMs. The code implementation for Section~\ref{sec:one-shot_method} and Section~\ref{sec:fine-tuning_method} is available in the associated repository\footnote{Code in the Sci-Sentence Repository – \url{https://github.com/fcobolanos/Classifying-the-Components-of-a-Literature-Review/tree/main/code}}.


\subsection{Zero-Shot Learning Settings}\label{sec:one-shot_method}

For the ZSL experiments, we evaluated a total of 12 decoder-only models. We excluded encoder-only and encoder-decoder architectures because they are generally not well-suited for ZSL tasks.

Among the 12 models, 8 were open-source and varied in size from 2 billion to 123 billion parameters. The remaining 4 were proprietary models whose parameter counts are not publicly available but are widely believed to exceed 700 billion. 
To ensure a fair comparison, we applied the same prompt template across all models. This template, refined through iterative prompt engineering following best practices~\citep{berryman2024prompt}, consists of three components: 1) an objective description to help the language model understand the task; 2) a precise explanation of the seven classification categories; and 3) a detailed procedure that includes the sentence to be classified along with a clearly specified output format to facilitate automated parsing. 
For transparency and reproducibility, the full prompt template is provided in Appendix~\ref{secB}.


The LLMs were executed using three different systems: 1) Amazon Bedrock\footnote{Amazon Bedrock – \url{https://aws.amazon.com/bedrock/}}, 2) the OpenAI API\footnote{OpenAI API – \url{https://openai.com/api/}}, and 3) KoboldAI\footnote{KoboldAI – \url{https://github.com/KoboldAI/KoboldAI-Client}}. Amazon Bedrock provides a pre-configured generic wrapper that standardises interactions with various supported LLMs, such as the MistralAI family, Anthropic models, and Llama models. The OpenAI API was used to access ChatGPT. KoboldAI, an open-source tool based on llama.cpp, enables the local execution of LLMs and exposes them via an API endpoint. In our setup, we used KoboldAI to load quantised models on a Google Colaboratory instance equipped with Nvidia V100 and L4 GPUs. The parameter configurations of all models are detailed in Appendix~\ref{secC}.

\subsection{Fine-tuning Settings}\label{sec:fine-tuning_method}

For the fine-tuning experiments, we evaluated 25 models spanning decoder, encoder, and encoder-decoder architectures, aiming to identify the most effective approach for this task. Each model was fine-tuned twice: first using the base version of the Sci-Sentence benchmark and then using its augmented counterpart, in order to evaluate the impact of data augmentation.

We employed two distinct fine-tuning strategies, tailored to the specific architecture of each model. For encoder models, we converted the sentences and their corresponding categories into tensors (Appendix~\ref{secE}, Section~\ref{subsec:encoder}). For decoder models (Sections~\ref{subsec:decoder_small}, \ref{subsec:decoder_medium}, and \ref{subsec:decoder_large} of Appendix~\ref{secE}) and encoder-decoder models (Appendix~\ref{secE}, Section~\ref{subsec:encoder_decoder}), we constructed prompts in a manner similar to Zero-Shot Learning, but without including category examples.


The models were fine-tuned using Google Colaboratory instances equipped with Nvidia V100 and L4 GPUs. For GPT-4o-mini, however, we relied on the OpenAI API. To mitigate the computational and memory constraints of Google Colaboratory, we adopted QLoRA~\citep{dettmers2024qlora}, a method that quantizes model weights from high-precision (32-bit) to low-precision (8-bit) formats. This quantization significantly reduces both computational overhead and memory usage.

To further reduce the number of trainable parameters in our decoder models, we evaluated two optimisation techniques: Low-Rank Adaptation (LoRA)~\citep{hu2021lora} and Noisy Embedding Instruction Fine-Tuning (NEFT)~\citep{jain2023neftune}. LoRA introduces small, trainable matrices derived from a low-rank decomposition of weight updates. During inference, these updates are combined with the original weights to produce the final output. In contrast, NEFT adds random noise to embedding vectors during training. We selected LoRA due to its widespread adoption as a standard optimisation method~\citep{mao2024survey}, and NEFT for its demonstrated effectiveness in improving performance~\citep{zhao2024long,laurenccon2024matters,li2024eagle,pradeep2023rankzephyr}.

To accelerate the fine-tuning procedure, we employed Unsloth\footnote{Unsloth - \url{https://github.com/unslothai/unsloth?utm_source=chatgpt.com}} as it offers up to 30× faster training and a 90\% reduction in memory consumption without compromising accuracy. In cases where Unsloth was not applicable (e.g., unsupported models or decoder-small models not needing optimization), we used Hugging Face Transformers. Appendix~\ref{secD} provides comprehensive details on parameter values and platforms for each model.

In Section~\ref{sec:results} (Results), we will focus on the best configuration for each model in terms of training data (base vs. augmented) and optimisation technique (NEFT vs. LoRA). We report the complete set of results is available in the associated repository\footnote{Results in the Sci-Sentence Repository – \url{https://github.com/fcobolanos/Classifying-the-Components-of-a-Literature-Review/tree/main/results}}.

\subsection{Overview of the Models}\label{sec:list of models}

\begin{table}[]
\caption{Main characteristics of the selected LLMs. The models are grouped by architectural type. M = million, B = billion, and T = trillion. \textbf{Context} length is measured in number of tokens. Not discl. means information not disclosed.}
\footnotesize
\label{tab:llm_models}
\begin{tabular}{l|l|r|r|r}
\toprule
\textbf{Category}                     & \textbf{Model Name}                       & \textbf{\# Param.} & \textbf{Context} & \textbf{Trained}  \\
\toprule
\multirow{4}{*}{Encoder}         & SciBERT                          & 110 M        & 512      & 3.1 B    \\
                                 & BioBERT                          & 110 M        & 512      & 18 B     \\
                                 & BigBird                          & 25 M         & 4,096    & 1.5 B    \\
                                 & BERT                             & 110 M        & 512      & 3.3 B    \\
\midrule
\multirow{12}{*}{\begin{tabular}[c]{@{}l@{}}Decoder\\ ZSL\end{tabular}}    & Llama2 (Open-Source   Full)      & 70   B       & 4,096    & 2 T     \\
                                 & Llama3 8b (Open-Source Full)     & 8 B          & 8,000    & 15  T   \\
                                 & Llama3 70b (Open-Source Full)    & 70  B        & 8,192    & 15 T    \\
                                 & Mistral (Open-Source Full)       & 7 B          & 2,000    & Not discl.  \\
                                 & Mixtral (Open-Source Full)       & 46.7 B       & 32,000   & Not discl.  \\
                                 & Mistral Large (Open-Source Full) & 123 B        & 32,000   & Not discl.  \\
                                 & Gemma (Open-Source Quantised)    & 2 B          & 8,192    & 2 T     \\
                                 & Orca (Open-Source Quantised)     & 13 B         & 4,096    & Not discl.  \\
                                 & Sonnet (Proprietary)             & Not discl.      & 200,000  & Not discl.  \\
                                 & Haiku (Proprietary)              & Not discl.      & 200,000  & Not discl.  \\
                                 & GPT-3.5 (Proprietary)            & Not discl.      & 16,000   & Not discl.  \\
                                 & GPT-4 (Proprietary)              & Not discl.      & 128,000  & Not discl.  \\
\midrule
\multirow{7}{*}{\begin{tabular}[c]{@{}l@{}}Decoder\\Small-FT\end{tabular}}   & Gemma2-2B                        & 2 B          & 8,192    & 2 T     \\
                                 & Olmo-1B                          & 1 B          & 2,048    & 3 T     \\
                                 & SmolLM2                          & 1.7 B        & 2,048    & 11 T    \\
                                 & TinyLlama                        & 1.1 B        & 2,048    & 3 T     \\
                                 & Arcee-lite (Merged)                       & 1.5 B        & 32,000   & Not discl.  \\
                                 & Phi3.5                           & 3.8 B        & 128,000  & 3.4 T   \\
                                 & Llama3.2-3B                      & 3 B          & 8,000    & 15 T    \\
\midrule
\multirow{6}{*}{\begin{tabular}[c]{@{}l@{}}Decoder\\Medium-FT\end{tabular}}  & Nemotron-8B                      & 8 B          & 4,096    & 3.5 T   \\
                                 & Olmo-7B                          & 7 B          & 2,048    & 2.5 T   \\
                                 & Mistral-7                        & 7 B          & 2,000    & Not discl.  \\
                                 & SuperNova-Lite (Merged)                    & 8 B          & 128,000  & Not discl.  \\
                                 & Llama3-8B                        & 8 B          & 8,000    & 15  T   \\
                                 & Arcee-Spark (Merged)                       & 7 B          & 128,000  & Not discl.  \\
\midrule
\multirow{4}{*}{\begin{tabular}[c]{@{}l@{}}Decoder\\Large-FT\end{tabular}}   & GPT-4o-mini                      & Not discl.      & 28,000   & Not discl.  \\
                                 & SuperNova-Medius (Merged)                 & 14 B         & 31,072   & Not discl.  \\
                                 & Gemma2-9B                        & 9 B          & 8,192    & 2 T     \\
                                 & Mistral-Nemo                     & 12 B         & 128,000  & Not discl.  \\
\midrule
\multirow{4}{*}{\begin{tabular}[c]{@{}l@{}}Encoder\\Decoder\end{tabular}} & T5 xxl                           & 11 B         & 512      & Not discl.  \\
                                 & T5 Large                         & 770 M        & 512      & 34 B     \\
                                 & T5                               & 222 M        & 512      & 1 T     \\
                                 & T5 xl                            & 3 B          & 512      & Not discl. \\
\bottomrule
\end{tabular}
\end{table}

In summary, the experiments involved a total of 37 approaches, including 12 using zero-shot learning and 25 using fine-tuning. These approaches varied in training parameters, quantisation strategies, openness (i.e., open-source vs. proprietary), fine-tuning methods, and model architectures. This comprehensive analysis enabled us to evaluate not only the performance of individual models but also the effectiveness of specific architectures and optimisation techniques for the task at hand.


To facilitate the analysis of the results, we divided the models into six categories based on their architecture (encoder, decoder, or encoder-decoder), training setting (ZSL vs. fine-tuning), and model size. 
These categories are: \textit{Encoder} (4 models), \textit{Encoder-decoder} (4 models), \textit{Decoder-ZSL} (12 decoders in ZSL setting), \textit{Decoder-Small-FT} (7 fine-tuned decoders with fewer than 4B parameters), \textit{Decoder-Medium-FT} (6 fine-tuned decoders between 4B and 8B), and \textit{Decoder-Large-FT} (4 fine-tuned decoders with more than 8B parameters). 

Table~\ref{tab:llm_models} presents all the models grouped by category, including their number of parameters, context window size, and the size of the datasets used during their original pretraining. Additional details about each model are provided in Appendix~\ref{secE}.

With respect to model size, although the number of parameters is widely regarded in the literature as the standard metric, there is no clear consensus on the specific thresholds that distinguish small from large models~\citep{wang2024comprehensive}. For example, \cite{liu2024mobilellm} define small models as those containing approximately one billion parameters, while \cite{fu2023specializing} consider models with up to ten billion parameters to still fall within the small category. Due to this lack of agreement, we established our own thresholds to achieve a relatively balanced distribution of models across size categories.

Finally, we note that we also chose to include four merged models (Arcee-lite, Arcee-Spark, SuperNova-Lite and SuperNova-Medius) in our evaluation. These models are constructed by combining the weights or architectures of multiple pre-trained LLMs to leverage their complementary strengths. This model merging technique efficiently enhances LLMs by integrating specialized knowledge and capabilities from different models into a single, more robust, and adaptable system. In particular, Arcee Lite is derived from a Qwen2-based architecture and represents a distilled variant of the Phi-3-Medium model. Arcee-Spark is also derived from a Qwen2-based architecture, and distilled from the Qwen2-7B-Instruct model. The SuperNova-Lite model is constructed on the Llama-3.1-8B-Instruct24 architecture and results from the distillation of the more expansive Llama-3.1-405B-Instruct model. Furthermore, SuperNova-Medius employs the Qwen2.5-14B-Instruct architecture and incorporates distilled knowledge from both the Qwen2.5-72B-Instruct and Llama-3.1-405B-Instruct models.

\section{Results}\label{sec:results}


In this section, we discuss the results of our experiments. As described in Section~\ref{sec:methodology}, all models were evaluated on the test set of the Sci-Sentence Benchmark, and their performance was assessed using precision, recall, and F1-score.

The first three subsections present an overall analysis of model performance. Section~\ref{sec:one-shot_results} reports the results obtained by the 12 ZSL decoder models. Section~\ref{sec:fine-tuning_results} describes the best-performing configurations for each of the 25 fine-tuned models. Finally, Section~\ref{sec:zsl_fine-tuning_results} compares the top-performing models across the seven categories of the schema.  
The following three subsections (Sections \ref{subsec:error_analysis}–\ref{subsec:augmented_data}) provide an in-depth analysis to interpret and contextualise the results presented earlier. Specifically, Section \ref{subsec:error_analysis} details the most prevalent model errors. Section \ref{subsec:optimisation_techniques} investigates the effects of applying optimisation techniques (LoRA, NEFT) to the decoder models. Finally, Section \ref{subsec:augmented_data} evaluates the impact of data augmentation on model performance across different optimisation strategies and training data solutions (base vs. augmented).


In the following analysis, we clearly distinguish between proprietary models, which are beyond our control and may involve additional undocumented processing steps, and open models, which were executed entirely within our configuration. While proprietary models may achieve superior performance, they are also less replicable. Therefore, we consider it important to evaluate them in a separate category.



\subsection{Zero-Shot Learning}\label{sec:one-shot_results}

Table~\ref{tab:prompting_metrics} presents the performance of the 12 LLMs in ZSL for classifying each of the 7 categories, along with their average performance.

Sonnets achieves the highest overall F1 score (82.6\%), followed by GPT-4 (76.8\%) and Mistral Large (74.9\%). These results confirm that the largest LLMs are capable of performing well on this task, although there is still considerable room for improvement.

When focusing on the open models (the first eight columns), the Mistral AI family, including Mistral Large, Mistral, and Mixtral, exhibits strong performance, achieving average F1 scores of 74.9\%, 72.6\%, and 71.0\%, respectively. 
Notably, Llama 3 70B also achieves solid results (F1 score of 69.4\%) and particularly excels in precision, reaching values above 85.0\% in all categories except \cat{Result}.



Regarding the proprietary models, as previously mentioned, Sonnet achieves the highest F1-score, closely followed by GPT-4. Both models demonstrate a well-balanced performance in terms of precision and recall across all categories. 
However, for certain categories, both are actually outperformed by the best open Mistral models. In particular, Sonnet is surpassed by Mistral in the \cat{Result} category (65.6\% vs. 80.9\%), and by Mistral Large in the \cat{Limitation} category (64.5\% vs. 78.0\%). This suggests that, although these proprietary models perform best overall, they can still be challenged, and even outperformed, by open models in specific areas.



Finally, we can observe a recurring pattern in which the majority of the models exhibit low precision in the \cat{Result} category, as well as low recall in the \cat{Description} and \cat{Limitation} categories. We will discuss more specific error patterns in the following sections.

\begin{table}[!h]
\caption{Precision, Recall, and F1-score of experiments with Zero-Shot Learning. PR= Precision, RE=Recall, F1=F1-score.\label{tab:prompting_metrics}}
\scriptsize
\begin{tabular}{l|l|p{0.6cm}|p{0.6cm}|p{0.6cm}|p{0.6cm}|p{0.6cm}|p{0.6cm}|p{0.6cm}|p{0.6cm}||p{0.6cm}|p{0.6cm}|p{0.6cm}|p{0.6cm}}
\toprule
\multicolumn{2}{r|}{\textbf{MODEL}}                 & \Rot{\textbf{Llama2}} & \Rot{\textbf{Llama3 8b}} & \Rot{\textbf{Llama3 70b}} & \Rot{\textbf{Mistral}} & \Rot{\textbf{Mixtral}} & \Rot{\textbf{Mistral Large}} & \Rot{\textbf{Gemma}} & \Rottt{\textbf{Orca}}  & \Rot{\textbf{Sonnet}} & \Rot{\textbf{Haiku}} & \Rot{\textbf{GPT-3.5}} & \Rote{\textbf{GPT-4}} \\
\toprule
\multirow{5}{*}{PR} & Average      & 0.542   & 0.471     & \textbf{0.862}      & 0.760   & 0.737   & 0.797         & 0.181 & 0.678 & \textbf{0.898}  & 0.733 & 0.624   & 0.824 \\ \cmidrule(lr){2-14}
                           & Overall      & 0.600   & \textbf{1.000}     & 0.905      & 0.654   & 0.682   & 0.818         & 0.158 & 0.533 & \textbf{1.000}  & 0.750 & 0.611   & 0.800 \\
                           & Research Gap & 0.455   & \textbf{1.000}     & \textbf{1.000}      & 0.789   & 0.882   & 0.727         & 0.125 & 1.000 & 0.900  & 0.875 & \textbf{1.000}   & \textbf{1.000} \\
                           & Description  & 0.556   & 0.333     & \textbf{1.000}      & 0.833   & 0.538   & 0.857         & 0.125 & 0.275 & \textbf{1.000}  & 0.500 & 0.167   & 0.833 \\
                           & Result       & 0.404   & 0.204     & 0.377      & \textbf{0.773}   & 0.594   & 0.500         & 0.172 & 0.513 & 0.488  & 0.486 & 0.404   &\textbf{0.556} \\
                           & Limitation   & 0.174   & 0.244     & 0.900      & 0.739   & \textbf{0.909}   & 0.800         & 0.227 & 0.900 & \textbf{1.000}  & 0.722 & 0.667   & \textbf{1.000} \\
                           & Extension    & 0.667   & 0.000     & 0.850      & 0.640   & 0.682   & \textbf{0.875}         & 0.059 & 0.667 & \textbf{0.900}  & 0.800 & 0.516   & 0.809 \\
                           & Other        & 0.938   & 0.513     & \textbf{1.000}      & 0.895   & 0.870   & \textbf{1.000}         & 0.400 & 0.857 & \textbf{1.000}  & \textbf{1.000} & \textbf{1.000}   & 0.769 \\
\midrule
\multirow{5}{*}{RE}    & Average      & 0.477   & 0.363     & 0.713      & 0.736   & 0.720   & \textbf{0.751}         & 0.157 & 0.561 & \textbf{0.822}  & 0.707 & 0.573   & 0.770 \\ \cmidrule(lr){2-14}
                           & Overall      & 0.545   & 0.091     & \textbf{0.864}      & 0.773   & 0.682   & 0.818         & 0.136 & 0.364 & \textbf{0.818}  &\textbf{0.818} & 0.500   & 0.727 \\
                           & Research Gap & 0.526   & 0.474     & \textbf{0.842}      & 0.789   & 0.789   & \textbf{0.842}         & 0.158 & 0.526 & \textbf{0.947}  & 0.737 & 0.474   & 0.737 \\
                           & Description  & 0.278   & 0.056     & 0.056      & 0.278   & \textbf{0.389}   & 0.333         & 0.167 & 0.611 & \textbf{0.611}  & 0.222 & 0.056   & 0.556 \\
                           & Result       & 0.950   & 0.450     & \textbf{1.000}      & 0.850   & 0.950   & 0.850         & 0.250 & 1.000 & \textbf{1.000}  & 0.900 & 0.950   & \textbf{1.000} \\
                           & Limitation   & 0.191   & 0.524     & 0.429      & \textbf{0.809}   & 0.476   & 0.762         & 0.238 & 0.429 & 0.476  & \textbf{0.619} & 0.381   & 0.524 \\
                           & Extension    & 0.100   & 0.000     & \textbf{0.850}      & 0.800   & 0.750   & 0.700         & 0.050 & 0.400 & \textbf{0.900}  & 0.800 & 0.800   & 0.850 \\
                           & Other        & 0.750   & 0.950     & 0.950      & 0.850   & \textbf{1.000}   & 0.950         & 0.100 & 0.600 & \textbf{1.000}  & 0.850 & 0.850   & \textbf{1.000} \\
\midrule
\multirow{5}{*}{F1}  & Average      & 0.455   & 0.312     & 0.694      & 0.726   & 0.710   & \textbf{0.749}         & 0.154 & 0.567 & \textbf{0.826}  & 0.701 & 0.553   & 0.768 \\ \cmidrule(lr){2-14}
                           & Overall      & 0.571   & 0.167     & \textbf{0.884}      & 0.708   & 0.682   & 0.818         & 0.146 & 0.432 & \textbf{0.900}  & 0.783 & 0.550   & 0.762 \\
                           & Research Gap & 0.488   & 0.643     & \textbf{0.914}      & 0.789   & 0.833   & 0.780         & 0.140 & 0.690 & \textbf{0.923}  & 0.800 & 0.643   & 0.849 \\
                           & Description  & 0.370   & 0.095     & 0.105      & 0.417   & 0.452   & \textbf{0.480}         & 0.143 & 0.379 & \textbf{0.759}  & 0.308 & 0.083   & 0.667 \\
                           & Result       & 0.567   & 0.281     & 0.548      & \textbf{0.809}   & 0.731   & 0.630         & 0.204 & 0.678 & 0.656  & 0.632 & 0.567   &\textbf{0.714} \\
                           & Limitation   & 0.182   & 0.333     & 0.581      & 0.773   & 0.625   & \textbf{0.780}         & 0.233 & 0.581 & 0.645  & 0.667 & 0.485   & \textbf{0.688} \\
                           & Extension    & 0.174   & 0.000     & \textbf{0.850}      & 0.711   & 0.714   & 0.778         & 0.054 & 0.500 & \textbf{0.900}  & 0.800 & 0.627   & 0.829 \\
                           & Other        & 0.833   & 0.667     & \textbf{0.974}      & 0.872   & 0.930   & \textbf{0.974}         & 0.160 & 0.706 & \textbf{1.000}  & 0.919 & 0.919   & 0.870\\
\bottomrule
\end{tabular}
\end{table}

\subsection{Fine-tuned Models}\label{sec:fine-tuning_results}

The fine-tuning experiments evaluated 25 models, grouped into the categories previously introduced in Section~\ref{sec:methodology}: \textit{Encoder} (4 models), \textit{Encoder-decoder} (4 models), \textit{Decoder-Small-FT} (7 models), \textit{Decoder-Medium-FT} (6 models), and \textit{Decoder-Large-FT} (4 models). Note that \textit{Decoder-ZST} is excluded here, as it was analysed in the preceding subsection.

As discussed in Section~\ref{sec:fine-tuning_method}, the fine-tuning experiments were performed using a range of configurations. In particular, each decoder was fine-tuned under four different settings, obtained by combining two training sets (base and augmented) with two optimisation methods (LoRA and NEFT). In contrast, encoders and encoder-decoder models were fine-tuned on both training sets using the standard fine-tuning procedure. To ensure clarity and avoid unnecessary detail, we report only the best-performing configuration for each model. A more detailed analysis of the effects of the optimisation techniques and training sets on performance is presented in Section~\ref{subsec:optimisation_techniques}, and Section~\ref{subsec:augmented_data}.


Table~\ref{tab:ft_overall_results} presents the F1-score, precision, and recall of the fine-tuned models. Within each architectural type, the models are listed in order of decreasing F1-score. In this section, we discuss their average performance across all rethorical categories, while the next section examines the performance of the top models within each category.


\begin{table}[!h]
\caption{Ranking of models by type. In \textbf{Conf}iguration: \textbf{B}=Base version of benchmark, \textbf{A}=Augmented benchmark, \textbf{L}=LoRA, \textbf{N}=NEFT.}
\footnotesize
\label{tab:ft_overall_results}
\begin{tabular}{l|l|r|r|r|r}

\toprule
 \multirow{2}{*}{\textbf{Model Type}}      & \multirow{2}{*}{\textbf{Model Name}}      &   \multirow{2}{*}{\textbf{Conf.}}            &  \multirow{2}{*}{\textbf{Precision}} &  \multirow{2}{*}{\textbf{Recall}}  & \multirow{2}{*}{\textbf{F1-Score}} \\
             \textbf{ }                    &             &                                 &    &  & 
                                   \\
\toprule
\multirow{4}{*}{Encoder}         & SciBERT   &  A                     & 0.929     & 0.928   & \textbf{0.928}    \\
                                 & BioBERT   &  A                         & 0.881     & 0.882   & 0.878    \\
                                 & BigBird   &  A                          & 0.886     & 0.881   & 0.878    \\
                                 & BERT   &  A                       & 0.871     & 0.866   & 0.861    \\
\midrule
\multirow{7}{*}{Decoder-Small}   & Gemma2-2B   &  AL         & 0.931     & 0.930    & \textbf{0.928}    \\
                                 & Olmo-1B   &  BN                               & 0.926     & 0.924   & 0.921    \\
                                 & SmolLM2   &  AN                     & 0.923     & 0.916   & 0.914    \\
                                 & TinyLlama   &  AN                   & 0.891     & 0.879   & 0.879    \\
                                 & Arcee-lite   &  BL                            & 0.887     & 0.882   & 0.878    \\
                                 & Phi3.5   &  BN                       & 0.872     & 0.870    & 0.869    \\
                                 & Llama3.2-3B   &  BL                      & 0.859     & 0.856   & 0.857    \\
\midrule                   
                                
\multirow{6}{*}{Decoder-Medium}  & Nemotron-8B   &  AL           & 0.940      & 0.937   & \textbf{0.936}    \\
                                 & Olmo-7B   &  AN                     & 0.938     & 0.938   & 0.935    \\
                                 & Mistral-7   &  BL                      & 0.937     & 0.934   & 0.933    \\
                                 & SuperNova-Lite   &  BL              & 0.932     & 0.927   & 0.928    \\
                                 & Llama3-8B   &  BN                        & 0.919     & 0.917   & 0.914    \\
                                 & Arcee-Spark   &  BN                           & 0.890      & 0.887   & 0.886    \\
\midrule
\multirow{4}{*}{Decoder-Large}   & gpt-4o-mini   &  B                         & 0.966     & 0.963   & \textbf{0.964}    \\
                                 & SuperNova-Medius   &  AL        & 0.945     & 0.943   & 0.943    \\
                                 & Gemma2-9B   &  BL                   & 0.943     & 0.943   & 0.942    \\
                                 & Mistral-Nemo   &  AL & 0.933     & 0.927   & 0.929    \\
\midrule
\multirow{4}{*}{Encoder-Decoder} & T5 xxl    &  B                                         & 0.910      & 0.898   & \textbf{0.899}    \\
                                 & T5 Large   &  A                         & 0.893     & 0.894   & 0.892    \\
                                 & T5   &  A                         & 0.882     & 0.882   & 0.879    \\
                                 & T5 xl   &  A                            & 0.860      & 0.858   & 0.856   
                                    \\
\bottomrule
  \end{tabular}
\end{table}

The results are clearly superior to those obtained in the ZSL setting, confirming the value of the Sci-Sentence Benchmark training sets in enabling high-performance models for this task.

In terms of model type, decoder-based models outperform encoder-based ones, which in turn outperform encoder-decoder architectures. Among the decoder models, model size plays a significant role. \textit{Decoder-Large-FT} achieved the best results, with GPT-4 reaching the highest F1-score of 96.4\%, followed by \textit{Decoder-Medium-FT} and \textit{Decoder-Small-FT}. 
Notably, SuperNova-Medius, a merged open-source model with 14B parameters trained on the augmented dataset, obtained the second-best result overall, with an F1-score of 94.3\%. 
This also represents a 17.5 percentage point improvement in F1-score over GPT-4 in the ZSL setting, which is commonly adopted as a default in many corporate solutions. These findings validate the importance of the novel Sci-Sentence datasets and demonstrate that open models can be highly competitive.

In the \textit{Decoder-Large-FT} category, GPT-4o-mini achieved the highest F1-score (96.4\%), followed by SuperNova-Medius (94.3\%) and Gemma2-9B (94.2\%). Within the \textit{Decoder-Medium-FT} category, the top-performing models showed very similar results. Nemotron-8B achieved the best score (93.6\%), followed closely by Olmo-7B (93.5\%) and Mistral-7B (93.3\%). The \textit{Decoder-Small-FT} models performed only slightly worse than the medium-sized decoders, with Gemma2-2B reaching the highest score in this category (92.8\%).

Notably, the \textit{Encoder} category produced results comparable to the smaller decoder models. In particular, Sci-BERT, a BERT variant pre-trained on academic text and therefore well-suited for this task, achieved an F1-score of 92.8\%, matching the performance of Gemma2-2B. This outcome is especially interesting because encoder models are generally faster and more scalable than small decoders. Thus, they offer an efficient solution for sentence classification with only a 1.5 percentage point drop in F1-score compared to the best-performing open model (SuperNova-Medius), and a 3.6 point drop compared to the top proprietary solution (GPT-4o-mini).

The encoder-decoder models did not perform particularly well, with the exception of T5-XXL, which achieved a solid 89.9\% F1. This result suggests that this architecture may not be particularly well suited to the task.

Regarding the merged models, while SuperNova-Medius achieved an excellent result as the first open model, the other two merged models did not perform as well. SuperNova-Lite matched SciBERT’s F1-score of 92.8\%, whereas Arcee-Lite and Arcee-Spark reported lower F1-scores, both falling below 90\%.


\subsection{Performance across rhetorical categories}\label{sec:zsl_fine-tuning_results}


To assess how the categories in the annotation schema influence the performance of different classifier architectures, Table~\ref{tab:ft_best_models} reports the F1-score, precision, and recall of the best models for each architecture type (encoder, encoder–decoder, and small, medium, and large decoder) across the seven categories of the annotation schema. The table also includes the performance of the ZSL decoders (D-ZSL).

\begin{table}[!h]
\caption{\textbf{PR}ecision, \textbf{RE}call, and \textbf{F1}-score of the best performing models by architectural type. In configuration: \textbf{B}=Base version of benchmark, \textbf{A}=Augmented benchmark, \textbf{L}=LoRA, \textbf{N}=NEFT.}
\footnotesize
\label{tab:ft_best_models}
\begin{tabular}{l|l|r|r|r|r|r|r|r}
\toprule
\multicolumn{2}{l|}{\textbf{Model Type}}            & \textbf{Enc}    & \textbf{D-ZSL} & \textbf{D-Sma}    & \textbf{D-Med}     & \textbf{D-Lar} & \textbf{Enc-Dec} & \textbf{} \\
\midrule
\multicolumn{2}{r|}{\textbf{MODEL}}            & \Rot{\textbf{SciBERT(A)}} & \Rot{\textbf{Sonnet}}   & \Rot{\textbf{Gemma2-2B(L)}} & \Rot{\textbf{Nemotron-8B(L)}} & \Rot{\textbf{GPT-4o-mini}} & \Rot{\textbf{T5 xxl}} & \Rote{\textbf{Average}} \\
\midrule
\multicolumn{2}{r|}{\textbf{Configuration}} & A & B & BL & BL & B & B & - \\ \midrule
\multirow{8}{*}{PR} & Average      & 0.929       & 0.898             & 0.931          & 0.940           &\textbf{ 0.966}       & 0.910          & 0.929 \\  \cmidrule(lr){2-9}
                           & Overall      & \textbf{1.000}       & \textbf{1.000}             & 0.955          & 0.950           & \textbf{1.000}       & 0.800          & 0.951 \\
                           & Research Gap & 0.857       & 0.900             & 0.864          & 0.905           & \textbf{0.950}       & 0.864          & 0.890 \\
                           & Description  & 0.941       & \textbf{1.000}             & 0.895          & 0.889           & \textbf{1.000}       & \textbf{1.000}         & 0.954 \\
                           & Result       & 0.895       & 0.488             & 0.947          & \textbf{1.000}           & 0.905       & 0.809          & 0.841 \\
                           & Limitation   & 0.857       & \textbf{1.000}             & \textbf{1.000}          & \textbf{1.000}           & \textbf{1.000}       & 0.947          & 0.967 \\
                           & Extension    & \textbf{0.952}       & 0.900             & 0.857          & 0.833           & 0.909       & 0.950          & 0.900 \\
                           & Other        & \textbf{1.000}       & \textbf{1.000}             & \textbf{1.000}          & \textbf{1.000}           & \textbf{1.000}       & \textbf{1.000}         & 1.000 \\
\midrule
\multirow{8}{*}{RE}    & Average      & 0.928       & 0.822             & 0.930          & 0.937           & \textbf{0.963}       & 0.898          & 0.913 \\ \cmidrule(lr){2-9}
                           & Overall      & 0.955       & 0.818             & 0.955          & 0.864           & \textbf{1.000}       & 0.909          & 0.917 \\
                           & Research Gap & 0.947       & 0.947             & \textbf{1.000}          & \textbf{1.000}           & \textbf{1.000}       & \textbf{1.000}         & 0.982 \\
                           & Description  & 0.889       & 0.611             & \textbf{0.944}          & 0.889           & 0.889       & 0.722          & 0.824 \\
                           & Result       & 0.850       & \textbf{1.000}             & 0.900          & 0.900           & 0.950       & 0.850          & 0.908 \\
                           & Limitation   & 0.857       & 0.476             & 0.809          & \textbf{0.905}           & \textbf{0.905}       & 0.857          & 0.802 \\
                           & Extension    & \textbf{1.000}       & 0.900             & 0.900          & \textbf{1.000}           & \textbf{1.000}       & 0.950          & 0.958 \\
                           & Other        & \textbf{1.000}       & \textbf{1.000}             & \textbf{1.000}          & \textbf{1.000}           & \textbf{1.000}       & \textbf{1.000}         & 1.000 \\
\midrule
\multirow{8}{*}{F1}  & Average      & 0.928       & 0.826             & 0.928          & 0.936           & \textbf{0.964}       & 0.899          & 0.914 \\ \cmidrule(lr){2-9}
                           & Overall      & 0.977       & 0.900             & 0.955          & 0.905           & \textbf{1.000}       & 0.851          & 0.931 \\
                           & Research Gap & 0.900       & 0.923             & 0.927          & 0.950           & \textbf{0.974}       & 0.927          & 0.934 \\
                           & Description  & 0.914       & 0.759             & 0.919          & 0.889           & \textbf{0.941}       & 0.839          & 0.877 \\
                           & Result       & 0.872       & 0.656             & 0.923          & \textbf{0.947}           & 0.927       & 0.829          & 0.859 \\
                           & Limitation   & 0.857       & 0.645             & 0.895          & \textbf{0.950}           & \textbf{0.950}       & 0.900          & 0.866 \\
                           & Extension    & \textbf{0.976}       & 0.900             & 0.878          & 0.909           & 0.952       & 0.950          & 0.928 \\
                           & Other        & \textbf{1.000}       & \textbf{1.000 }            & \textbf{1.000}          & \textbf{1.000}           & \textbf{1.000}       & \textbf{1.000} & 1.000 \\
\bottomrule                           
\end{tabular}
\end{table}

GPT-4o-mini and Nemotron-8B, the top-performing large and medium-sized models respectively, achieve the highest overall performance, with average F1-scores of 96.4\% and 93.6\%. In contrast, Sonnet, in the zero-shot learning (ZSL) setting, records the lowest average F1-score among the best models by type, underperforming by 13.8 and 11.0 percentage points compared to GPT-4o-mini and Nemotron-8B, respectively.

Notably, GPT-4o-mini achieves the highest absolute F1-score in five of the seven categories. Nemotron-8B matches GPT-4o-mini in the \cat{Limitation} category and achieves the best result in \cat{Result}. Finally, and perhaps surprisingly, the smaller SciBERT yields the best performance in \cat{Extension}.

Focusing on category-specific performance, \cat{Other} is clearly the easiest category to identify. All top models achieve perfect scores in this case. A closer examination suggests this is because classification errors tend to occur among semantically similar categories, such as confusing \cat{Description} or \cat{Limitation} with \cat{Result}, whereas \cat{Other} is semantically distinct enough to be reliably recognised by most systems.

By contrast, the categories \cat{Limitation} and \cat{Description} are the most challenging to classify. This is particularly evident from the performance of Sonnet in the ZSL setting, where it achieves a Recall below 61\% for both categories. The category \cat{Result} also yields a low F1-score, but for a different reason. Although it is identified with high accuracy, leading to a high Recall, it exhibits a relatively low Precision.

However, models fine-tuned on Sci-Sentence demonstrate a substantially better understanding. In particular, Nemotron-8B achieves F1-scores of 94.7\% and 95.0\% on \cat{Limitation} and \cat{Result}, respectively. A similar, though less pronounced, trend is observed for \cat{Description}, where the best ZSL method attains an F1-score of 75.9\%, while the best fine-tuned model reaches 94.1\%. 

In summary, we identify four key insights. 
First, the current generation of LLMs can perform exceptionally well on this task when fine-tuned on high-quality datasets, such as Sci-Sentence, achieving performance levels exceeding 96\%. In contrast, ZSL produces substantially lower results, underscoring the critical role of domain-specific training data. 
Second, while large proprietary models such as GPT-4o attain the highest performance, lightweight open-source alternatives, such as SuperNova-Medius and Nemotron-8B, also yield excellent results. These models offer additional benefits in terms of scalability, reproducibility, and transparency. 
Third, although decoder-only models achieve the best overall performance, encoder-based models pre-trained on relevant data, such as SciBERT, can still deliver very competitive results. Moreover, they offer significantly higher scalability, making them a practical alternative when processing large volumes of text. 
Finally, certain categories, notably \cat{Limitation} and  \cat{Description} remain particularly challenging, especially in ZSL settings. However, the use of high-quality training data enables satisfactory performance even in these more difficult cases.

\vspace{0.5cm}
\subsection{Error Analysis through Confusion Matrices}\label{subsec:error_analysis}

\begin{figure}[!ht]
    \begin{subfigure}[b]{0.47\textwidth}   
      \includegraphics[width=\textwidth]{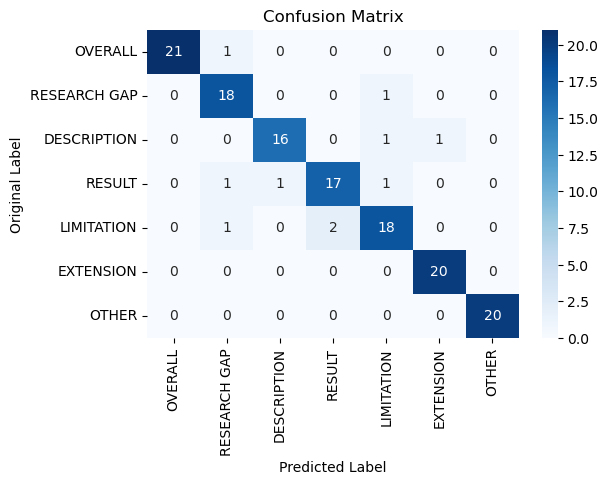}
      \caption{Encoder: SciBERT (Augmented)}
    \end{subfigure}
    \hfill
    \begin{subfigure}[b]{0.47\textwidth}
      \includegraphics[width=\textwidth]{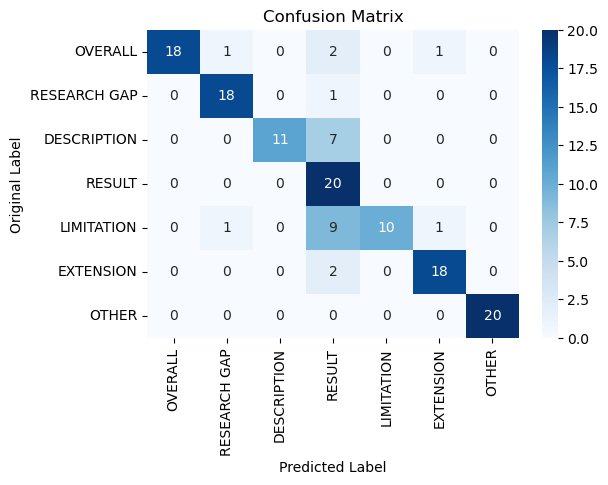}
      \caption{Decoder-ZSL: Sonnet}
    \end{subfigure}
    \begin{subfigure}[b]{0.47\textwidth} 
       \includegraphics[width=\textwidth]{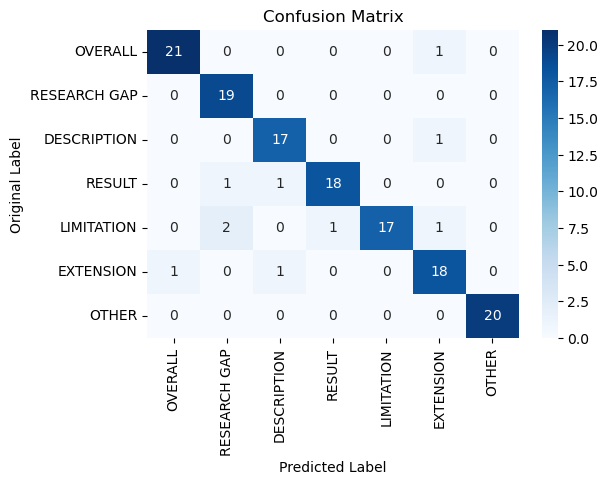}
      \caption{Encoder Small: Gemma2-2B (LoRA-Augmented)}
    \end{subfigure}
    \hfill
    \begin{subfigure}[b]{0.47\textwidth}
      \includegraphics[width=\textwidth]{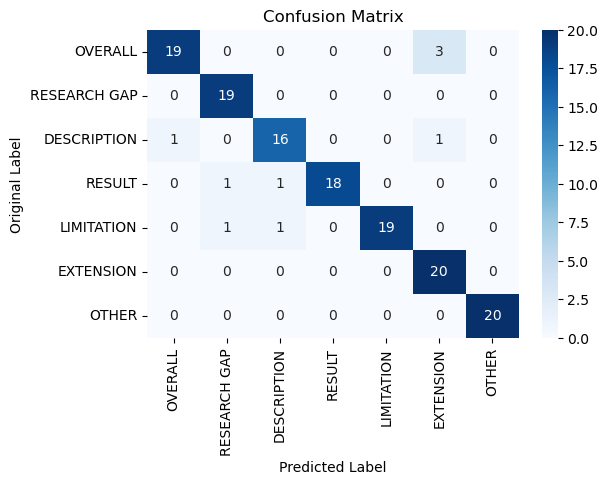}
      \caption{Decoder Medium: Nemotron-8B (LoRA Augmented)}
    \end{subfigure}
    \begin{subfigure}[b]{0.47\textwidth} 
       \includegraphics[width=\textwidth]{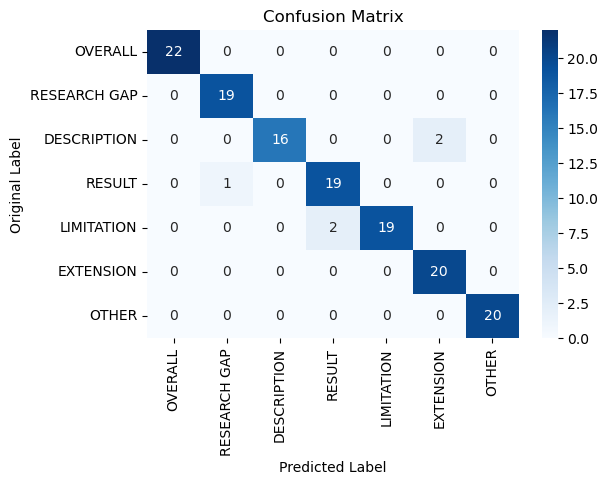}
      \caption{Encoder Large: GPT4o-mini}
    \end{subfigure}
    \hfill
    \begin{subfigure}[b]{0.47\textwidth}
      \includegraphics[width=\textwidth]{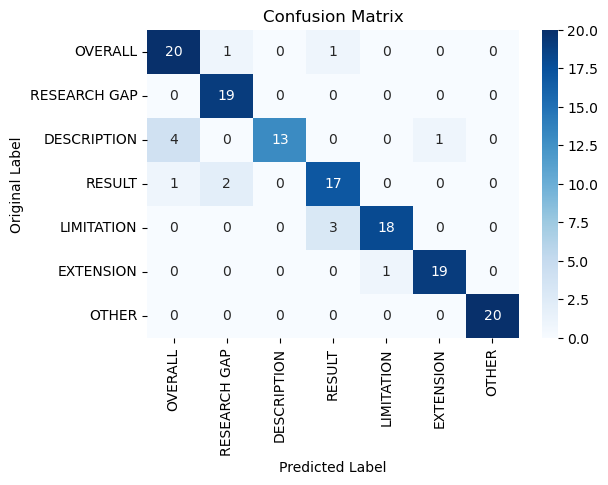}
      \caption{Encoder-Decoder: T5xxl}
    \end{subfigure}
    
    \caption{Confusion matrices of the best performing models by model type.}
    \label{confusion_matrics_fine-tuning}
  \end{figure}

Figure~\ref{confusion_matrics_fine-tuning} presents the confusion matrices for the best-performing model of each of the six architectural types, providing a more detailed view of the results reported in Table~\ref{tab:ft_best_models}. Among these, GPT-4o-mini achieved the highest performance, with only five misclassifications. In contrast, Sonnet in ZSL, which was the least accurate among the top models, recorded 25 misclassifications. The remaining models had error counts ranging from nine to fourteen.

The majority of misclassification occurred within the \cat{Limitation} and  \cat{Description} categories. 
Specifically, for the \cat{Limitation}  category, misclassification rates were 14\% for SciBERT, 52\% for Sonnet, 19\% for Gemma2-2B, 10\% for Nemotron3-8B, 10\% for GPT-4o-mini, and 14\% for T5. The \cat{Description} category showed misclassification rates of 11\% (SciBERT), 39\% (Sonnet), 11\% (Nemotron3-8B), 11\% (GPT-4o-mini), and 28\% (T5).

Sentences annotated as \cat{Limitation} in the gold standard were most often misclassified as \cat{Research Gap} or \cat{Result}.

A detailed analysis of these cases suggests that this occurs because instances of \cat{Limitation} are often phrased in a way that implies a broader lack of knowledge. This phrasing can lead them to be interpreted as a \cat{Research Gap}.
For example, consider the sentence: \textit{“Although the accuracy of current shape segmentation exceeds 90\%, some models might still be poorly segmented and thus fail to be accurately landmarked”}. The intended meaning is to describe a limitation of the specific model being examined. However, the generalized expression "current shape segmentation" implies a shortcoming affecting the entire field rather than a precise constraint of the study. As a consequence, such statements may be misclassified as a \cat{Research Gap}, since they seem to highlight an area where further research or improvement is needed across existing approaches instead of pointing to a limitation of the work at hand.


Moreover, some \cat{Limitation} sentences explicitly refer to the results, which can lead to their misclassification as belonging to the \cat{Result} category. For instance, the sentence \textit{“Our encoding of traces does allow for uneven data samples, but it is probable that the performance will be reduced when sampling becomes irregular or infrequent”} can be misclassified as a \cat{Result}, because it reports a finding about the model’s ability to handle uneven data. However, it simultaneously conveys a limitation by acknowledging that performance may deteriorate under certain conditions, thus blurring the distinction between reporting results and discussing constraints. This characteristic of sentences belonging to multiple rhetorical types is a specific challenge that we aim to address in future refinement efforts.

On the other hand, sentences annotated as \cat{Description} were most frequently misclassified as \cat{Overall} or \cat{Extension}. Our manual analysis indicates that some \cat{Description} sentences can be confused with introductory clauses that appear to summarise the general status of a method or topic. This can lead to their incorrect assignment to the \cat{Overall} category. This issue is exemplified by the sentence \textit{“Specifically, as the first inductive kg embedding method, mean [8] learns to represent entities using their neighbors by simply mean-pooling the information of neighboring entity-relation pairs".} The introductory phrase presents the method in a broad and contextual manner, highlighting its novelty or general contribution rather than providing a precise descriptive detail. This framing causes the sentence to resemble a high-level summary, which can lead to its misclassification as \cat{Overall}. However, the latter part of the sentence (\textit{“learns to represent entities using their neighbors…”}) in fact specifies how the method operates, which is characteristic of the \cat{Description} category.

In other cases, \cat{Description} outline the study's design in a way that resembles an expansion of previous methodology, aligning with the definition of the \cat{Extension} category. For example, the sentence \textit{“Telm [42] improves on TComplEx by utilizing a linear temporal regularizer and multi-vector embeddings to perform 4th-order tensor factorization of TKGs”} might appear to fit the \cat{Extension} category because it references a previous method. However, it should be classified as \cat{Description}, since it does not explain how the current study elaborates on or extends the previous one. For a sentence to qualify as \cat{Extension}, it must detail the nature of the elaboration, for example, by clarifying what limitations are addressed or what specific aspects of the prior study are developed further.

\subsection{Comparing Optimisation Techniques}\label{subsec:optimisation_techniques}

As discussed in Section~\ref{sec:fine-tuning_method}, the models were fine-tuned using two well-known techniques: LoRA and NEFT. In the previous sections, we always referred to the best-performing model between the two. However, it is also worth analysing in which cases one approach outperformed the other on Sci-Sentence benchmark.

Table~\ref{tab:ft_optimisation} compares the F1 scores of the models trained with LoRA and NEFT. The ``Diff.'' column reports the difference between the F1 score of the model using NEFT and that of the model using LoRA. Therefore, a positive value indicates that NEFT outperforms LoRA, whereas a negative value indicates that LoRA outperforms NEFT.

For large decoder models, LoRA consistently achieved slightly better results than NEFT. In the other categories, the outcomes are more mixed. Among medium decoder models, half of the models (Olmo 7B, Arcee-Spark, Llama3-8B) obtained F1-score improvements with NEFT, ranging from 0.019 to 0.029. For small decoder models, NEFT improved performance in three out of seven cases (Olmo 1B, TinyLlama, and Phi3.5), with TinyLlama exhibiting a notable F1-score increase of 0.150.

In conclusion, the evidence is insufficient to definitively establish the superiority of one optimization technique over the other for small and medium models. However, for large decoder models, LoRA produced more favourable results.

\begin{table}[!h]
\caption{Comparison of F1-scores when employing LORA or NEFT as optimisation techniques for optimizing decoder models.}
\footnotesize
\label{tab:ft_optimisation}
\begin{tabular}{l|l|l|l|l}
\toprule
\textbf{Model   Type}                    & \textbf{Model Name}                 & \textbf{F1 (LORA)} & \textbf{F1 (NEFT)} & \textbf{Diff.} \\
\midrule
\multirow{7}{*}{Decoder-Small}  & Olmo-1B                    & 0.902   & 0.921   & \textbf{0.019}      \\
                                & TinyLlama                  & 0.552   & 0.702   & \textbf{0.150}      \\
                                & Arcee-lite                 & 0.878   & 0.863   & -0.015     \\
                                & SmolLM2                    & 0.796   & 0.782   & -0.014     \\
                                & Gemma2-2B        & 0.876   & 0.815   & -0.061     \\
                                & Llama3.2-3B        & 0.857   & 0.843   & -0.014     \\
                                & Phi3.5            & 0.861   & 0.869   & \textbf{0.008}      \\
\midrule
\multirow{6}{*}{Decoder-Medium} & Olmo-7B                    & 0.900     & 0.929   & \textbf{0.029}      \\
                                & Mistral-7B        & 0.933   & 0.891   & -0.042     \\
                                & Arcee-Spark                & 0.872   & 0.891   & \textbf{0.019}      \\
                                & Llama3-8B          & 0.892   & 0.914   & \textbf{0.022}      \\
                                & Llama-3.1-SuperNova-Lite   & 0.928   & 0.922   & -0.006     \\
                                & Nemotron-8B       & 0.885   & 0.835   & -0.050     \\
\midrule                                
\multirow{3}{*}{Decoder-Large}  & Gemma2-9B        & 0.942   & 0.891   & -0.051     \\
                                & Mistral-Nemo & 0.907   & 0.89    & -0.017     \\
                                & SuperNova-Medius       & 0.902   & 0.87    & -0.032 \\
\midrule
\end{tabular}
\end{table}

\subsection{Assessing the Efficacy of Synthetic Data}\label{subsec:augmented_data}

Semi-synthetic training data produced by LLMs have proven to be very effective, but their performance is inconsistent across different classification tasks~\citep{li2023synthetic}. One of our goals was to determine whether semi-synthetic data within Sci-Sentence, generated by creating alternative versions of manually classified sentences from the original data, could improve performance.  

To this end, we trained each model on both the augmented and the original training data.  
As for the  optimization techniques, in the previous section we always considered the best-performing model between the two configurations.  
Table~\ref{tab:ft_augmented} compares the F1 scores of the models trained on the augmented data and on the original data. The \textit{Performance Gain} row refers to the difference between the F1 score obtained with the augmented data and the F1 score obtained with the original data. A positive value, therefore, indicates an improvement in performance due to the augmented data. It is important to note that Sonnet~3.0, which was used to generate the augmented datasets, was not included among the fine-tuning models in order to avoid potential issues of circularity~\citep{clarke2024llm}.

\begin{table}[!h]
\caption{Comparison of F1-score of models trained on the Base or Augmented Benchmark. L=LoRA, N=NEFT.}
\footnotesize
\label{tab:ft_augmented}
\begin{tabular}{l|l|r|r|r}
\toprule

\textbf{Archit.}      & \multirow{2}{*}{\textbf{Model Name}}      &   \textbf{F1-score} & \textbf{F1-score}  & \textbf{Performance} \\
\textbf{Type} &     & \textbf{Base}   & \textbf{Augmented} &        \textbf{Gain}  \\
\midrule
\multirow{4}{*}{Encoder}         & BERT              & 0.590             & 0.861        & \textbf{0.271}      \\
                                 & SciBERT         & 0.870             & 0.928        & \textbf{0.058}      \\
                                 & BioBERT                 & 0.731            & 0.878        & \textbf{0.147}      \\
                                 & BigBird                 & 0.646            & 0.878        & \textbf{0.232}      \\
\midrule
\multirow{7}{*}{\begin{tabular}[c]{@{}l@{}}Decoder\\Small\end{tabular}}   & Olmo-1B [N]                      & 0.921            & 0.912        & -0.009     \\
                                 & TinyLlama [N]                  & 0.702            & 0.879        & \textbf{0.177}      \\
                                 & Arcee-Lite [L]                          & 0.878            & 0.863        & -0.015     \\
                                 & SmolLM2 [N]                    & 0.782            & 0.914        & \textbf{0.132}      \\
                                 & Gemma2-2B [L]        & 0.876            & 0.928        & \textbf{0.052}      \\
                                 & Llama3.2-3B [L]        & 0.857            & 0.852        & -0.005     \\
                                 & Phi3.5 [N]            & 0.869            & 0.831        & -0.038     \\
\midrule
\multirow{6}{*}{\begin{tabular}[c]{@{}l@{}}Decoder\\Medium\end{tabular}}  & Olmo-7B [N]                      & 0.929            & 0.935        & \textbf{0.006}      \\
                                 & Mistral-7B [L]           & 0.933            & 0.893        & -0.040      \\
                                 & Arcee-Spark [N]                & 0.886            & 0.806        & -0.080      \\
                                 & Llama3-8 [N]             & 0.914            & 0.901        & -0.013     \\
                                 & Llama-3.1-SuperNova-Lite [L]   & 0.928            & 0.927        & -0.001     \\
                                 & Nemotron-8B [L]          & 0.885            & 0.936        & \textbf{0.051}      \\
\midrule
\multirow{4}{*}{\begin{tabular}[c]{@{}l@{}}Decoder\\Large\end{tabular}}   & Gemma2-9B [L]          & 0.942            & 0.929        & -0.013     \\
                                 & Mistral-Nemo [L] & 0.907            & 0.929        & \textbf{0.022}      \\
                                 & SuperNova-Medius [L]       & 0.902            & 0.943        & \textbf{0.041}      \\
                                 & GPT-4o-mini             & 0.964            & 0.892        & -0.072     \\
\midrule
\multirow{4}{*}{\begin{tabular}[c]{@{}l@{}}Encoder\\Decoder\end{tabular}} & T5                 & 0.849            & 0.879        & \textbf{0.030}       \\
                                 & T5   Large               & 0.875            & 0.892        & \textbf{0.017}      \\
                                 & T5 xl                    & 0.839            & 0.856        & \textbf{0.017}      \\
                                 & T5 xxl                   & 0.899            & 0.892        & -0.007    \\
\midrule
\end{tabular}
\end{table}

An interesting insight is that encoder models benefit the most from augmented data, with all models improving in performance, sometimes very significantly. For example, the original BERT achieves a gain of over 27 percentage points in F1 score. Even a domain-specific model such as SciBERT shows a clear improvement, increasing from 87.0\% to 92.8\% thanks to the augmented data. This finding has important implications, particularly for applications that require a lightweight model to scalably annotate a large number of research papers, where a lightweight encoder model may therefore be preferred.

Encoder-decoder models also tend to benefit from augmented data, especially in their smaller variants, but not to the same extent as encoder models. In sum, our results of semi-synthetic data increasing the performance of encoders or encoder-decoder architectures align with existing  literature~\citep{fields2024survey,chae2023large,fatemi2024comparative}.

For decoder models, the performance gains are more variable. This observation is consistent with the literature: while some studies report a positive effect~\citep{singh2023beyond,li2024aide,zhezherau2024hybrid,chen2024diversity}, others find little or no benefit~\citep{guo2023curious,zhao2024understanding,li2024gradual,mecklenburg2024injecting,maheshwari2024efficacy}. The greatest benefits are observed in the smaller variants: three out of seven (namely TinyLlama, SmolLM2, and Gemma2-2B) show notable improvements, with TinyLlama achieving a substantial F1-score increase of over 30\%. Among the medium-sized models, only two exhibit improvements. It is interesting to note that one of them is Nemotron-8B, which is also the best-performing model in this category.

For the large models, two out of four (Mistral-Nemo and SuperNova-Medius) demonstrate enhanced performance. Notably, SuperNova-Medius achieves a significant improvement of 4.1 percentage points, making it the best open model among all those tested. Conversely, GPT-4o-mini does not benefits from augmented data. 

In conclusion, while the improvements in decoder models are not consistent across all cases, those that do benefit tend to gain a substantial margin, allowing them to outperform other open alternatives in the same category. Indeed, in all categories, the best-performing open model was trained on augmented data. We can therefore conclude that augmented data can be highly beneficial for decoder models as well, although careful consideration is required to identify which decoders are most likely to benefit.

\section{Conclusion}\label{sec:conlusion}

In this paper, we propose a novel framework for classifying sentences in the related work or literature review sections of research papers into seven categories, extending previous research in this area. The goal is to develop an automated method for identifying sentences that present research gaps, limitations, extensions of previous work, and similar aspects, in order to support advanced retrieval-based systems for question answering and literature review generation. We conduct a comprehensive evaluation of a wide range of encoder, encoder-decoder, and decoder language models with different architectures on this task. To facilitate this evaluation, we create and publicly release the \textit{Sci-Sentence} benchmark, which includes a base version with 700 manually annotated sentences and an augmented version with a total of 2,940 sentences, combining both manually annotated and semi-synthetic samples.




These experiments provided several novel insights that significantly advance the state of the art in this space. 
First, the current generation of LLMs can perform remarkably well on this task when fine-tuned on high-quality datasets such as Sci-Sentence, achieving performance levels above 96\% F1. 
Second, although large proprietary models such as GPT-4o achieve the highest performance, lightweight open-source alternatives, including SuperNova-Medius and Nemotron-8B, also deliver excellent results.
Third, while decoder-only models attain the best overall performance, small and scalable encoder-based models pre-trained on relevant data, such as SciBERT, remain highly competitive. This makes them a practical choice for processing large volumes of text efficiently.
Finally, augmenting the original data with semi-synthetic examples generated by LLMs for fine-tuning has proven effective, particularly by enabling small encoders to achieve robust results and substantially improving the performance of several open decoders. \color{black} 
In summary, the proposed framework, the Sci-Sentence benchmark, and our experimental results together constitute an important step and a foundational contribution to the ``Related Work Generation'' task.

It is important to acknowledge a few limitations of this study, which we aim to address in future work. First, our dataset was predominantly drawn from Computer Science, so further investigation is needed to assess the generalisability of these findings to other fields. Second, the field of Artificial Intelligence is rapidly evolving, with newer LLMs being released in recent weeks. These advancements could lead to more efficient and effective classifications. However, we believe that the fundamental insights emerging from this analysis are unlikely to change in the medium term.
Finally, additional research is required to enhance classifiers’ ability to distinguish the most complex and challenging categories, such as \cat{Description} and \cat{Result}.

As future work, we plan to advance our research on multiple fronts. First, we aim to extend the classifier from single-label to multi-label in order to better capture the multifaceted nature of complex sentences that may pertain to multiple categories.  
Second, we plan to investigate how to capture more elusive categories, such as critiques and interpretations of a piece of literature, which are crucial for incorporating critical evaluations, nuanced perspectives, and broader contextual understanding into our representation. Although existing work, such as \cite{khoo2011analysis}, identifies an author’s `interpretation' category, we argue that a more fine-grained approach is necessary to effectively support the automatic generation of literature reviews, instead of aggregating all interpretations into a single category.
Finally, we intend to develop a novel framework for generating automatic literature reviews that integrates the framework presented in this paper and moves beyond uncritical multi-document summarisation towards producing high-quality, in-depth analyses of the literature.

\begin{appendices}
\newpage
\section{Augmented Data}\label{secA}

\begin{figure}[!h]
  \centering
  \includegraphics[width=0.8\textwidth]{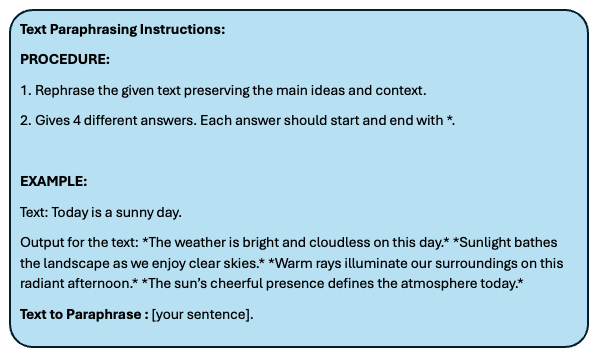}
  \caption{Prompt used to generate semi-synthetic data.}
  \label{fig:prompt_synthetic}
\end{figure}

In Figure~\ref{fig:prompt_synthetic}, we present the prompt used to generate the semi-synthetic data. Table~\ref{tab:Syntactic Similarity} reports the average syntactic similarity for the training and validation sets. For each generated sentence, we calculate its Levenshtein distance to the original sentence, and the average Levenshtein distance to all other generated sentences.

\begin{table}[!h]
\caption{Average syntactic similarity for the training and validation sets. Original refers to the original sentence. Other refers to the other synthetic sentences. Syn=Synthetic.  }
\scriptsize
\centering
\label{tab:Syntactic Similarity}
\begin{tabular}
{p{1.8cm}|p{1.8cm}|p{.5cm}|p{.50cm}|p{.50cm}|p{.50cm}|p{.50cm}|p{.50cm}|p{.50cm}|p{.50cm}}
\toprule
\textbf{Data Type}                 & \textbf{Category}     &  \Rot{\textbf{Syn1-Original}} &  \Rot{\begin{tabular}{@{}c@{}}\textbf{Syn1-Other} \\ \textbf{Averaged}\end{tabular}} &  \Rot{\textbf{Syn2-Original}} &  \Rot{\begin{tabular}{@{}c@{}}\textbf{Syn2-Other} \\ \textbf{Averaged}\end{tabular}} &  \Rot{\textbf{Syn3-Original}} &  \Rot{\begin{tabular}{@{}c@{}}\textbf{Syn3-Other} \\ \textbf{Averaged}\end{tabular}} &  \Rot{\textbf{Syn4-Original}} &  \Rote{\begin{tabular}{@{}c@{}}\textbf{Syn4-Other} \\ \textbf{Averaged}\end{tabular}} \\
\midrule
\multirow{8}{*}{Training}   & Average      & 0.57                                      & 0.56                                    & 0.54                                      & 0.57                                    & 0.54                                      & 0.56                                    & 0.53                                       & 0.55                                    \\
                            & Overall      & 0.57                                      & 0.54                                    & 0.54                                      & 0.55                                    & 0.54                                      & 0.55                                    & 0.53                                       & 0.53                                    \\
                            & Research Gap & 0.61                                      & 0.61                                    & 0.55                                      & 0.60                                    & 0.53                                      & 0.59                                    & 0.54                                       & 0.58                                    \\
                            & Description  & 0.51                                      & 0.53                                    & 0.51                                      & 0.54                                    & 0.53                                      & 0.54                                    & 0.52                                       & 0.54                                    \\
                            & Result       & 0.60                                      & 0.56                                    & 0.57                                      & 0.57                                    & 0.55                                      & 0.56                                    & 0.56                                       & 0.54                                    \\
                            & Limitation   & 0.60                                      & 0.59                                    & 0.55                                      & 0.60                                    & 0.57                                      & 0.60                                    & 0.55                                       & 0.56                                    \\
                            & Extension    & 0.54                                      & 0.55                                    & 0.53                                      & 0.55                                    & 0.50                                      & 0.55                                    & 0.51                                       & 0.52                                    \\
                            & Other        & 0.59                                      & 0.57                                    & 0.54                                      & 0.57                                    & 0.53                                      & 0.57                                    & 0.54                                       & 0.56                                    \\
\midrule
\multirow{8}{*}{Validation} & Average      & 0.57                                      & 0.56                                    & 0.54                                      & 0.56                                    & 0.52                                      & 0.55                                    & 0.54                                       & 0.54                                    \\
                            & Overall      & 0.54                                      & 0.54                                    & 0.53                                      & 0.54                                    & 0.50                                      & 0.52                                    & 0.50                                       & 0.50                                    \\
                            & Research Gap & 0.66                                      & 0.63                                    & 0.62                                      & 0.62                                    & 0.61                                      & 0.58                                    & 0.58                                       & 0.60                                    \\
                            & Description  & 0.47                                      & 0.50                                    & 0.49                                      & 0.54                                    & 0.45                                      & 0.51                                    & 0.48                                       & 0.53                                    \\
                            & Result       & 0.52                                      & 0.50                                    & 0.50                                      & 0.52                                    & 0.46                                      & 0.49                                    & 0.46                                       & 0.50                                    \\
                            & Limitation   & 0.68                                      & 0.67                                    & 0.57                                      & 0.64                                    & 0.63                                      & 0.67                                    & 0.64                                       & 0.63                                    \\
                            & Extension    & 0.54                                      & 0.52                                    & 0.48                                      & 0.52                                    & 0.48                                      & 0.52                                    & 0.57                                       & 0.52                                    \\
                            & Other        & 0.58                                      & 0.55                                    & 0.56                                      & 0.55                                    & 0.50                                      & 0.54                                    & 0.53                                       & 0.53    \\
\bottomrule
\end{tabular}
\end{table}

\clearpage

\section{Prompt for Zero-Shot Learning}\label{secB}

\begin{figure}[!h]
  \centering
  \includegraphics[width=0.8\textwidth]{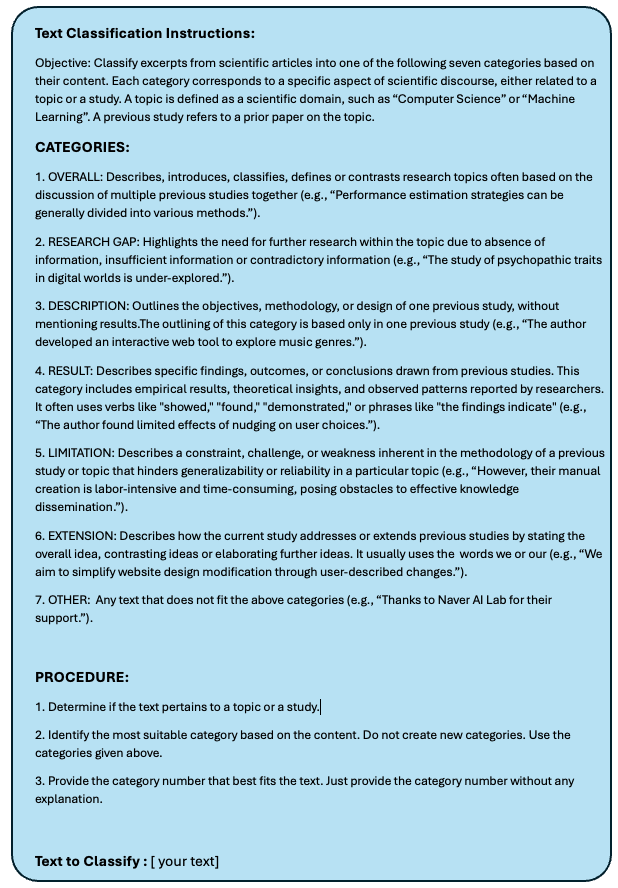}
  \caption{Prompt used in the Zero-Shot Learning experiments.}
  \label{fig:prompt_zsl}
\end{figure}

Figure~\ref{fig:prompt_zsl} shows the prompt used in the Zero-Shot Learning experiments.

\clearpage

\section{Parameters Settings for  Zero-Shot Learning}\label{secC}

Table~\ref{tab:SettingsZSL} summarizes the settings used in the Zero-Shot Learning experiments.

\begin{table}[!h]
\centering
\scriptsize
\caption{Zero-Short Learning Settings.}
\label{tab:SettingsZSL}
\begin{tabular}{l|l|l|l|l}
\toprule
\textbf{Model}     & \textbf{Temperature} & \textbf{Top\_k} & \textbf{Top\_p} & \textbf{Platform}  \\   
\midrule
Llama 2       & 0           & NA     & 0      & Amazon Bedrock \\
Llama 3 8b    & 0           & NA     & 0      & Amazon Bedrock \\
Llama 3 70b   & 0           & 1      & 0      & Amazon Bedrock \\
Mistral       & 0           & 1      & 0      & Amazon Bedrock \\
Mixtral       & 0           & 1      & 0      & Amazon Bedrock \\
Mistral Large & 0           & 1      & 0      & Amazon Bedrock \\
Gemma         & 0           & 1      & 0      & KoboldAI       \\
Orca          & 0           & 1      & 0      & KoboldAI       \\
Sonnet        & 0           & 1      & 0      & Amazon Bedrock \\
Haiku         & 0           & 1      & 0      & Amazon Bedrock \\
GPT-3.5       & 0           & 1      & 0      & OpenAI API     \\
GPT-4         & 0           & 1      & 0      & OpenAI API \\
\bottomrule
\end{tabular}
\end{table}

\section{Parameters Settings for Fine-Tuning}\label{secD}

Table~\ref{tab:SettingsFinetuning} reports the configuration settings employed for fine-tuning the models.

\begin{table}[!h]
\centering
\scriptsize
\caption{Fine-tuning Settings.}
\label{tab:SettingsFinetuning}
\begin{tabular}{l|l|l|l|l|l|l}
\toprule
\multirow{2}{*}{\textbf{Model}}   & \multicolumn{3}{l}{\textbf{LORA}} & \textbf{NEFT}  & \multirow{2}{*}{\textbf{\# Epochs}} & \multirow{2}{*}{\textbf{Platform}} \\
                         & \textbf{r}     & \textbf{alpha} & \textbf{droput}  & \textbf{alpha} &                                             \\
\midrule
Olmo-1B                  & 256   & 128    & 0.1     & 5     & 4                          & Hugging Face              \\
TinyLlama                & 256   & 128    & 0.1     & 5     & 1                          & Hugging Face              \\
Arcee-lite               & 256   & 128    & 0.05    & 5     & 1                          & Hugging Face              \\
SmolLM2                  & 256   & 128    & 0.1     & 5     & 1                          & Hugging Face              \\
Gemma-2-2b               & 256   & 128    & 0.1     & 5     & 1                          & Unsloth                   \\
Llama3.2                 & 256   & 128    & 0       & 5     & 1                          & Unsloth                   \\
Phi3.5                   & 256   & 128    & 0       & 5     & 1                          & Unsloth                   \\
Olmo-7B                  & 16    & 32     & 0.1     & 5     & 4                          & Hugging Face              \\
Mistral-7b               & 16    & 15     & 0       & 5     & 4                          & Unsloth                   \\
Arcee-Spark              & 256   & 128    & 0.1     & 5     & 1                          & Hugging Face              \\
Llama3-8b                & 256   & 128    & 0.1     & 5     & 1                          & Unsloth                   \\
Llama-3.1-SuperNova-Lite & 256   & 128    & 0       & 5     & 1                          & Unsloth                   \\
Nemotron-8B              & 16    & 32     & 0.1     & 5     & 1                          & Hugging Face              \\
Gemma-2-9b               & 256   & 128    & 0.1     & 5     & 1                          & Unsloth                   \\
Mistral-Nemo             & 16    & 16     & 0       & 5     & 4                          & Unsloth                   \\
SuperNova-Medius-14B     & 256   & 128    & 0.1     & 5     & 1                          & Hugging Face              \\
gpt-4o-mini-2024-07-18   & NA    & NA     & NA      & NA    & 1                          & OpenAI API\\
\bottomrule
\end{tabular}
\end{table}

\clearpage

\section{List of models}\label{secE}

This appendix outlines the complete list of LLMs tested in our experiments, grouped into seven categories: Decoder-Zero Shot Learning, Decoder-Small, Decoder-Medium, Decoder-Large, Encoder-Decoder, and Encoder.

\subsection{Decoder-Zero Shot Learning }\label{subsec:decoder-osl}

\subsubsection{Full Models}

\textbf{Llama 2 Chat 70B\footnote{ Llama 2 Chat 70B - \url{(https://llama.meta.com/llama2/)}}} (shortened as llama2) has been trained using a dataset comprising 2 trillion tokens derived from publicly accessible online sources. It has  70 billion parameters  and a context length of 4,096 tokens. 

\textbf{Llama 3 8B Instruct\footnote{ Llama 3 8b Instruct - \url{(https://ai.meta.com/blog/meta-llama-3/)}}} (shortened as llama3 8B full) is an auto-regressive language model that employs an optimised transformer architecture with 8 billion parameters. It possesses a context length of 8,000 tokens and has been trained on a dataset consisting of 15 trillion tokens. The training process involved supervised fine-tuning (SFT) in conjunction with reinforcement learning from human feedback (RLHF) to align the model with human preferences for utility and safety. To enhance inference scalability, the model incorporates Grouped-Query Attention (GQA). 

\textbf{Llama 3 70b Instruct\footnote{Llama 3 70b Instruct - \url{(https://ai.meta.com/blog/meta-llama-3/)}}} (shortened as llama3 70b) is composed of 70 billion parameters  and a context length of 8,192 tokens. This model has been fine-tuned and optimised specifically for dialogue and chat use cases based.  

\textbf{Mistral 7b Instruct~\citep{jiang2023mistral}} (shortened as mistral)
is a 7 billion parameter model with a context length of 32,000 tokens. It incorporates architectural innovations such as Sliding Window Attention mechanism, GQA, and Byte-fallback Byte Pair Encoding (BPE) tokenizer. The first architectural innovation, accommodates a context length of 8,000 tokens and features a fixed cache size, which theoretically enables it to process up to 128,000 tokens. The second innovation, enhances inference speed while reducing cache size. While the third innovation, ensures reliable character recognition without the need for out-of-vocabulary tokens. 

\textbf{Mixtral 8X7b Instruct~\citep{jiang2024mixtral}} (shortened as mixtral) consists of 46.7 billion parameters and is capable of processing a context length of 32,000 tokens. It is based on a Sparse Mixture of Experts (SMoE) architecture with open weights. This model employs the same innovative architecture as the Mistral 7b Instruct. However, the Sliding Window Attention mechanism restricts it to handling a context length of 8,000 tokens.  

\textbf{Mistral Large\footnote{ Mistral Large - \url{https://mistral.ai/news/mistral-large/}}} features a context length of 32,000 tokens and employs 123 billion parameters. The model was trained using a heterogeneous dataset that encompassed a substantial amount of code, multilingual information, and content across a broad spectrum of topics.

\subsubsection{Quantised Models}

\textbf{Gemma2-2B-Instruct\footnote{ Gemma2-2B-Instruct- \url{https://huggingface.co/google/gemma-2b-it-GGUF}}}(shortened as Gemma) is a quantized version of Google's Gemma-2b-it language model~\citep{team2024gemma} converted to the GPT-Generated Unified Format (GGUF) file format with 2 billion parameters and  context length of 8,192 tokens.  

\textbf{
Orca-2-13B\footnote{ 
Orca-2-13B - \url{https://huggingface.co/TheBloke/Orca-2-13B-GGUF}}}(shortened as Orca)
It is a quantized version of Microsoft's Orca 2~\citep{mukherjee2023orca}, converted to the GGUF format having 13 billion parameters and a context length of 4096 tokens~\citep{mitra2023orca}. It was fine-tuned on LLama 2 13B base model.

\subsubsection{Proprietary Models}

The precise specifications of these models are confidential. Consequently, we can only provide limited information about them. In particular, details about their parameters are not disclosed.

\textbf{Sonnet 3.0} (Sonnet) and \textbf{Haiku 3.0} (Haiku) are part of the Claude 3 series, a family of LLMs developed by Anthropic~\citep{anthropic2024claude}. Both models were trained on a proprietary corpus derived from both public and private sources. They have a context window of 200,000 tokens. Their distinction resides in the fact that Haiku is designed for immediate responses, while Sonnet 3.0 possesses the capability to manage complex tasks thanks to its architectural framework. These models adhered to the Constitutional AI framework, ensuring their alignment with the principles of helpfulness, honesty, and non-harmfulness

\textbf{GPT-3.5 Turbo} (GPT-3.5)  and \textbf{GPT-4 Turbo} (GPT-4), both developed by OpenAI, exhibit specific distinctions in their design and capabilities~\citep{openai2303gpt}. GPT-3.5 Turbo operates with a context window of 16,000 tokens, whereas GPT-4 Turbo features a larger context window of 128,000 tokens. Furthermore, GPT-4 Turbo is capable of processing both textual and visual inputs, whereas GPT-3.5 Turbo is restricted to textual data exclusively. Additionally, GPT-4 was trained on a larger and more heterogeneous dataset compared to  GPT-3.5 Turbo. Despite this, GPT-3.5 Turbo continues to be a practical and economical choice for numerous applications because of its effective combination of performance and efficiency.

\subsection{Decoder-Smal FT}\label{subsec:decoder_small}

The number of parameters in this category ranges from 1 billion to 3.8 billion. However, the context window and the number of training tokens differ across models. For instance, Olmo-1B-Instruct \textbf{(Olmo-1B)}~\citep{groeneveld2024olmo}, with 1 billion parameters, and TinyLlama-1.1B-Chat-v1.0 \textbf{(TinyLlama)}~\citep{zhang2024tinyllama}, with 1.1 billion parameters, share the same context window size of 2,048 tokens and were trained on 3 trillion tokens. In contrast, \textbf{SmolLM2}\footnote{ SmolLM2 - \url{https://github.com/huggingface/smollm/blob/main/README.md}}, which has 1.7 billion parameters, also employs a context window of 2,048 tokens but was trained on a substantially larger dataset of 11 trillion tokens. These models also vary in their training sources: Olmo-1B was trained on subsets of Dolma v1.7~\citep{soldaini2024dolma}, TinyLlama utilized the architecture and tokenizer of Llama 2~\citep{Touvron2023Llama2O}, and SmolLM2 was trained on a diverse dataset that includes textbooks, web content, code, mathematics, and external data sources.

On the other hand, Llama3.2-3-Instruct \textbf{(Llama3.2-3B)}\footnote{ Llama3.2-3B-Instruct - \url{https://huggingface.co/meta-llama/Llama-3.2-3B-Instruct}}  was trained on a dataset comprising 15 trillion tokens, with a parameter count of 3 billion and a context window extending up to 8,000 tokens. In contrast, Phi3.5-mini-Instruct \textbf{(Phi3.5)}\footnote{Phi3.5-mini-Instruct - \url{https://huggingface.co/microsoft/Phi-3.5-mini-instruct}} was trained on 3.4 trillion tokens, incorporating 3.8 billion parameters and a significantly larger context window of 128,000 tokens. Gemma2-2B-Instruct \textbf{(Gemma2-2B)} shares the same characteristics as those described in the category decoder OSL, differing only in its data source\footnote{ Gemma2-2B-Google - \url{https://huggingface.co/google/gemma-2-2b-it}}. The only merge model in this category is \textbf{Arcee-Lite}\footnote{ Arcee-Lite - \url{https://huggingface.co/arcee-ai/arcee-lite}} which have 1.5 billion parameters and was developed using Distilkit\footnote{ Distilkit - \url{https://github.com/arcee-ai/DistillKit}}. It supports a context window of 32,000 tokens and its distillation source is Phi-3-Medium~\citep{abdin2024phi}.

\subsection{Decoder-Medium FT}\label{subsec:decoder_medium}

In this category the models have 7 billion or 8 billion parameters. For instance, Nemotron 3-8B-chat \textbf{(Nemotron3-8B)}\footnote{Nemotron3-8B - \url{https://tinyurl.com/mw959vux}} has 8 billion parameters  and a context windows of 4,096 tokens. It was trained on 3.5 trillion tokens based on a large corpus of internet-scale data, including 53 languages and 37 coding languages. Similarly, models such as Mistral-7B-Instruct \textbf{(Mistral-7B)}, with 7 billion parameters, and Llama3-8-Instruct \textbf{(Llama3-8B)}, with 8 billion parameters, exhibit comparable features to their full-version counterparts but are distinguished by their quantized configurations. In the case of Olmo-7B-Instruct \textbf{(Olmo-7B)}, which also has 7 billion parameters, its distinction from Olmo-1B lies in its training dataset size of 2.5 trillion tokens.

For merged models, this category includes \textbf{Arcee-Spark}\footnote{Arcee-Spark - \url{https://huggingface.co/arcee-ai/Arcee-Spark}}  and Llama-3.1-SuperNova-Lite \textbf{(SuperNova-Lite)}\footnote{SuperNova-Lite - \url{https://huggingface.co/arcee-ai/Llama-3.1-SuperNova-Lite}}. \textbf{Arcee-Spark}, initialized from the Qwen2-7B-Instruct~\citep{yang2024qwen2}, comprises 7 billion parameters. On the other hand, \textbf{SuperNova-Lite}, with 8 billion parameters, is built upon the Llama-3.1-8B-Instruct\footnote{Llama-3.1 - \url{https://ai.meta.com/blog/meta-llama-3-1/}} architecture and distilled from the Llama-3.1-405B-Instruct model. Both models have a context window of 128,000 tokens.







\subsection{Decoder-Large FT}\label{subsec:decoder_large}

This category includes models with parameters ranging from 9 billion to 14 billion. In this sense, Mistral-Nemo-Instruct-2407 \textbf{(Mistral-Nemo)}\footnote{ Mistral Nemo - \url{https://mistral.ai/news/mistral-nemo/}} has 12 billion parameters and a 128,000 token context window. Its training dataset incorporates a mix of multilingual text, code data, and conversational-style data to ensure high-quality input. Whereas, \textbf{SuperNova-Medius}\footnote{ Mistral Nemo - \url{https://huggingface.co/arcee-ai/SuperNova-Medius}} is a merged model of 14 billion parameters based on the Qwen2.5-14B-Instruct architecture. It combines knowledge from both the Qwen2.5-72B-Instruct model and the Llama-3.1-405B-Instruct model through distillation. It has a context windows of 131,072 tokens. In contrast, Gemma2-9-Instruct  \textbf{(Gemma2-9B)} has the same features as in Gemma, but differs on its source\footnote{ Gemma2-9B-Google - \url{https://huggingface.co/google/gemma-2-9b-it}} and the number of parameters which are 9 billion.  The only proprietary model is GPT-4o-mini-2024-07-18 \textbf{(GPT-4o-mini)}\footnote{ GPT-4o-mini - \url{https://openai.com/index/gpt-4o-mini-advancing-cost-efficient-intelligence/}} which is part of the GPT-4o family and is designed to be a cost-efficient, high-performance AI model. It has multimodal capabilities and its context window of 128,000 tokens. It is designed to replace GPT-3.5 Turbo in ChatGPT due to its improved performance and cost-efficiency for various AI applications.





\subsection{Encoder-Decoder}\label{subsec:encoder_decoder}

For this category, we employed different versions of the \textbf{T5}~\citep{raffel2020exploring}, including T5-base (222 million parameters), T5-large (770 million parameters), T5-xl (3 billion parameters), and T5-xxl (11 billion parameters). These models are built on a transformer architecture, wherein the encoder processes the input text, and the decoder generates the output text. T5 has been pretrained on the  C4 corpus, a large dataset of text and code, using both supervised and self-supervised training methods. It has a context window of 512 tokens. Our decision to use T5 was driven by its superior performance compared to other encoder-decoder models~\citep{garg2021news,sarrouti2022comparing,kementchedjhieva2023exploration}.

\subsection{Encoder}\label{subsec:encoder}
In this category, all models, with the exception of \textbf{BigBird}~\citep{zaheer2020big}, exhibit identical features, including a number of parameters of 110 million, a context length of 512 tokens, and the utilisation of a full attention mechanism. In contrast, BigBird distinguishes itself with 125 million parameters, an extended context length of 4,096 tokens, and the use of a sparse attention mechanism.

BERT base~\citep{devlin2018bert}, hereafter referred to as \textbf{BERT},  was trained on a dataset containing 3.3 billion tokens sourced from Wikipedia and the Google Books Corpus. SciBERT case (\textbf{SciBERT})~\citep{beltagy2019scibert} was developed using 1.14 million scientific articles from Semantic Scholar\footnote{ Semantic Scholar - \url{(https://www.semanticscholar.org/)}}, covering the biomedical and computer science domains, with a total of 3.1 billion tokens. \textbf{BioBERT}~\citep{lee2020biobert} was trained on an extensive collection of biomedical literature, including publications from PubMed\footnote{ PubMed - \url{(https://pubmed.ncbi.nlm.nih.gov/)}} and PMC\footnote{ PMC - \url{(https://pmc.ncbi.nlm.nih.gov/)}}, and employs WordPiece~\citep{wu2016google} tokenization to efficiently manage a large vocabulary. On the other hand, \textbf{BigBird} was trained on a dataset of 1.5 billion tokens drawn from the Books Corpus and Wikipedia.

\end{appendices}

\bibliography{bibliography}
\end{document}